\documentclass[smallcondensed]{svjour3}
\smartqed 
\usepackage[authoryear ]{natbib}
\usepackage{mathptmx} 
\usepackage[utf8]{inputenc} % utf8 encoding
\usepackage{xcolor}
\usepackage{todonotes}
\usepackage{multirow}
\usepackage{makecell}
\usepackage[parfill]{parskip}
\usepackage{lineno,hyperref}
\modulolinenumbers[5]

\begin{document}
\title{Survey on Evaluation Methods for Dialogue Systems}
%\tnotetext[t1]{This paper has been partially funded by the LIHLITH project supported by ERA-NET CHIST-ERA and Swiss National Science Foundation (20CH21\_174237)}

\author{
Jan Deriu\and
Alvaro Rodrigo\and
Arantxa Otegi\and
Guillermo Echegoyen\and
Sophie Rosset\and
Eneko Agirre\and
Mark Cieliebak
}

\institute{Jan Deriu \and Mark Cieliebak \at
              Zurich University of Applied Sciences (ZHAW) \\
              Steinberggasse 13, 8400 Winterthur, Switzerland \\
              Tel.: +41 (0) 58 934 47 65\\
              \email{\{deri, ciel\}@zhaw.ch}       
           \and
           Alvaro Rodrigo \and Guillermo Echegoyen \at
              NLP \& IRGroup, UNED \\
              C/Juan del Rosal 16, Madrid 28040, Spain\\
              \email{\{alvarory, gblanco\}@lsi.uned.es}
           \and
           Arantxa Otegi \and Eneko Agirre \at
              IXA NLP group , University of the Basque Country(UPV/EHU)\\
              Manuel Lardizabal 1 Donostia, Basque Country 20018, Spain\\
              \email{\{arantza.otegi, e.agirre\}@ehu.eus}
           \and
           Sophie Rosset \at
              Université Paris-Saclay, CNRS, LIMSI \\
              Campus Universitaire, Bât. 508, rue John von Neumann, 91405 Orsay Cedex, France \\
              \email{sophie.rosset@limsi.fr}           
}

\date{Received: date / Accepted: date}

\maketitle

\begin{abstract}
In this paper, we survey the methods and concepts developed for the evaluation of dialogue systems. Evaluation, in and of itself, is a crucial part during the development process. Often, dialogue systems are evaluated by means of human evaluations and questionnaires. However, this tends to be very cost- and time-intensive. Thus, much work has been put into finding methods which allow a reduction in involvement of human labour. In this survey, we present the main concepts and methods. For this, we differentiate between the various classes of dialogue systems (task-oriented, conversational, and question-answering dialogue systems). We cover each class by introducing the main technologies developed for the dialogue systems and then present the evaluation methods regarding that class. 
\end{abstract}

% !TEX root = survey_dialogue_evaluation.tex

\section{Introduction}
As the amount of digital data continuously grows, users demand technologies that offer quick access to such data. In fact, users rely on systems that support information search interactions such as Siri\footnote{https://www.apple.com/es/siri/}, Google Assistant\footnote{https://assistant.google.com/}, Amazon Alexa\footnote{https://www.amazon.com} or Microsoft XiaoIce~\citep{zhou2018design}, etc. These technologies, called Dialogue Systems (DS), allow the user to converse with a computer system using natural language. Dialogue Systems are applied to a variety of tasks, e.g.:
\begin{itemize}
\item Virtual Assistants aid users in everyday tasks, such as scheduling appointments. They usually operate on predefined actions which can be triggered by voice command. 
\item Information-seeking systems provide users with information about a question (e.g. the most suitable hotel in town). These questions also include factual questions as well as more complex questions. 
\item E-learning dialogue systems train students for various situations. For instance, they train the interaction with medical patients or train military personnel in questioning a witness.
\end{itemize}

One crucial step in the development of DS is evaluation. That is, to measure how well the DS is performing. However, evaluating a dialogue system can prove to be problematic because there are two important factors to be considered. Firstly, the definition of what constitutes a high-quality dialogue is not always clear and often depends on the application. Even if a definition is assumed, it is not always clear how to measure it. For instance, if we assume that a high-quality dialogue system is defined by its ability to respond with an appropriate utterance, it is not clear how to measure appropriateness or what appropriateness means for a particular system. Moreover, one might ask the users if the responses were appropriate, but as we will discuss below, user feedback might not always be reliable for a variety of reasons.

The second factor is that the evaluation of dialogue systems is very cost- and time-intensive. This is especially true when the evaluation is carried out by a user study, which requires careful preparation,  the need for inviting and compensating users for their participation. 

Over the past decades, many different evaluation methods have been proposed. The evaluation methods are closely tied to the characteristics of the dialogue system which they are aimed at evaluating. Thus, quality is defined in the context of the function which dialogue system is meant to fulfil. For instance, a system designed to answer questions will be evaluated on the basis of correctness, which is not necessarily a suitable metric for evaluating a conversational agent.

Most methods are aimed at automating the evaluation, or at least automating certain aspects of the evaluation. The goal of an evaluation method is to obtain automated and repeatable evaluation procedures that allow efficient comparisons in the quality of different dialogue strategies.

This survey is structured as follows; in the next section we give a general overview over the different classes of dialogue systems and their characteristics. We then introduce the evaluation task in greater detail, with an emphasis on the goals of an evaluation and the requirements on an evaluation metric. In Sections \ref{sec:eval_task}, \ref{sec:eval_non_task}, and \ref{sec:eva_qa_dialogue}, we introduce each dialogue system class (i.e. task-oriented systems, conversational agents and question answering dialogue systems). Thereafter, we give an overview of the characteristics, dialogue behaviour, and concepts behind the implementation methods of the various dialogue systems. Finally, we present the evaluation methods and the ideas behind them. Here, we set an emphasis the relationship between these methods and the dialogue system classes, including which aspects of the evaluation are automated. In Section \ref{sec:datasets}, we give a short overview of the relevant datasets and evaluation campaigns in the domain of dialogue systems. In Section \ref{sec:discussion}, we discuss the issues and challenges in devising automated evaluation methods and discuss the level of automation achieved.
%
%!TEX root = survey_dialogue_evaluation.tex

\section{A General Overview}
\subsection{Dialogue Systems}
    Dialogue Systems (DS) usually structure dialogues in \emph{turns}, each turn is defined by one or more \emph{utterances} from one speaker. Two consecutive turns between two different speakers is called an \emph{exchange}. Multiple exchanges constitute a \emph{dialogue}. Another different, but related view is to interpret each turn or each utterance as an action (more on this later). 

The main component of a dialogue system is the dialogue manager that defines the content of the next utterance and thus the behaviour of the dialogue system. There are many different approaches to design a dialogue manager, which are partly dictated by the application of the dialogue system. However, there are three broad classes of dialogue systems that we encounter in the literature: task-oriented systems, conversational agents and interactive question answering systems\footnote{In recent literature, the distinction is made only between the first two classes of dialogue systems \citep{serban2015surveyJournal,Chen:2017:SDS:3166054.3166058,jura2017speech_dial}. However, interactive question answering systems cannot be completely placed in either of the two categories. }. 

We identified the following characteristic features that help differentiate between the three different classes: whether the system is developed to solve a task, whether the dialogue follows a structure, whether the domain is restricted or open, whether the dialogue spans over multiple turns, whether the dialogues are long or rather efficient, who takes the initiative, and what interface is used (text, speech, multi-modal). Table \ref{tbl:ds_char} depicts the characteristics for each of the dialogue system classes. In this table, we can see the following main features for each class:
    \begin{itemize}
    \item Task-oriented systems are developed to help the user solve a specific task as efficiently as possible. The dialogues are characterized by following a clearly defined structure that is derived from the domain. The dialogues follow mixed initiative; both the user and the system can take the lead. Usually, the systems found in the literature are built for speech input and output. However, task-oriented systems in the domain of assisting users are built on multi-modal input and output. 
    \item Conversational agents display a more unstructured conversation, as their purpose is to have open-domain dialogues with no specific task to solve. Most of these systems are built to emulate social interactions, and thus longer dialogues are desired.
    \item Question Answering (QA) systems are built for the specific task of answering questions. The dialogues are not defined by a structure as with task-oriented systems, however, they mostly follow the question and answer style pattern. QA systems may be built for a specific domain, but may be also tilted towards more open domain questions. Usually, the domain is dictated by the underlying data, e.g. knowledge bases or text snippets from forums. Traditional QA systems work on a singe-turn interaction, however, there are systems that allow multiple turns to cover follow-up questions. The initiative is mostly done by the user, who asks questions. 
    \end{itemize}

\begin{table}[h!]
\centering
\begin{tabular}{l||lll}
                & Task-oriented DS    & Conversational Agents & Interactive QA         \\ \hline \hline
Task            & Yes - clear defined & No                    & Yes - answer questions \\
Dial. Structure & Highly structured     & Not structured        & No                     \\
Domain          & Restricted          & Mostly open domain    & Mixed                  \\
Turns           & Multi               & Multi                 & Single/Multi           \\
Length          & Short               & Long                  & -                      \\
Initiative      & Mixed/ system init  & mixed/user init       & user init              \\
Interface       & multi-modal         & multi-modal           & mostly text           
\end{tabular}
\caption{Characterizations of the different dialogue system types. }
\label{tbl:ds_char}
\end{table}

\subsection{Evaluation}
Evaluating dialogue systems is a challenging task and subject of much research. We define the goal of an evaluation method as having an automated, repeatable evaluation procedure with high correlation to human judgments, which is able to differentiate between various dialogue strategies and is able to explain which features of the dialogue systems are important. Thus, the following requirements can be stated:
\begin{itemize}
\item Automatic: in order to reduce the dependency on human labour, which is time- and cost-intensive as well as not necessarily repeatable, the evaluation method should be automated, or at least partially automated. 
\item Repeatable: the evaluation method should yield the same result if applied multiple times to the same dialogue system under the same circumstances. 
\item Correlated to human judgments: the procedure should yield ratings that correlate to human judgments. 
\item Differentiate between different dialogue systems: the evaluation procedure should be able to differentiate between different strategies. For instance, if one wants to test the effect of a \emph{barge-in} feature (i.e. allowing the user to interrupt the dialogue system), the evaluation procedure should be able to highlight the effects. 
\item Explainable: the method should give insights into which features of the dialogue system impact the quality of the dialogue and in which manner they do so. For instance, the methods should reveal that the automatic speech recognition system's \emph{word-error rate} has a high influence on the quality of the natural language understanding component, which in turn impacts the intent classification. 
\end{itemize}
In this survey, we focus on the efforts of automating the evaluation process. This is a very difficult, but crucial task, as human evaluations are cost- and time-intensive. Although much progress has been made in automating the evaluations of dialogue systems, the reliance on human evaluation is still present. Here, we give a condensed overview on the human-based evaluations used in the literature.
\paragraph{Human Evaluation.} 
There are various approaches to a human evaluation. The test subjects can take on two main roles: interacting with the system or rating a dialogue or utterance, or both. In the following, we differentiate among different types of user populations. Among each of the populations, the subjects can take on any of the two roles. 
\begin{itemize}
\item Lab experiments: Before crowdsourcing was popular, dialogue systems were evaluated in a lab environment. Users were invited to participate in the lab where they interacted with a dialogue system and subsequently filled a questionnaire. For instance, \cite{Young:2010:HIS:1621140.1621240} recruited 36 subjects, which were given instructions and presented with various scenarios. The subjects were asked to solve a task using a spoken dialogue system. Furthermore, a supervisor was present to guide the users. The lab environment is very controlled, which is not necessarily comparable to the real world \citep{black2011sds_challenge,SCHMITT201512}.
\item In-field experiments: Here, the evaluation is performed by collecting feedback from real users of the dialogue systems~\citep{lamel2000limsi}. For instance, for the Spoken Dialogue Challenge \citep{black2011sds_challenge}, the systems were developed to provide bus schedule information in Pittsburgh. The evaluation was performed by redirecting the evening calls to the dialogue systems and getting the user feedback at the end of the conversation. The Alexa Prize \footnote{\url{https://developer.amazon.com/alexaprize}} also followed the same strategy, i.e. it let real users interact with operational systems and gathered user feedback over a span of several months.
\item Crowdsourcing: Recently, human evaluation has shifted from a lab environment to using crowdsourcing platforms such as Amazon Mechanical Turk (AMT). These platforms provide large amounts of recruited users. \cite{jurc2011AMT} evaluate the validity of using crowdsourcing for evaluating dialogue systems, and their experiments suggest that using enough crowdsourced users, the quality of the evaluation is comparable to the lab conditions. Current research relies on crowdsourcing for human evaluation \citep{serban2017mrrnn,wen2017e2e_dialog}. 

Especially conversational dialogue systems are evaluated via crowdsourcing, where there are two main evaluation procedures: crowdworkers either talk to the system and rate the interaction or they are presented with a context from the test set and a response by the system, which they need to rate. In both settings, the crowdworkers are aksed to rate the system based on quality, fluency or appropriateness. Recently, \cite{adiwardana2020towards} introduced Sensibleness and Specificity Average (SSA), where humans rate the sensibleness and specificity of a response. These capture two aspects of human behaviour: making sense and being specific. A dialogue system can be sensible by responding with vague answers (e.g. ``I don't know"), whereas it is only specific if it takes the context into account.
\end{itemize}
Human based evaluation is difficult to set up and to carry out. Much care has to be taken in setting up the experiments; the users need to be properly instructed and the tasks need to be prepared so that the experiment reflects real-world conditions as closely as possible. Furthermore, one needs to take into account the high variability of user behaviour, which is present especially in crowdsourced environments. 
\paragraph{Automated Evaluation Procedures.}
A procedure which satisfies the aforementioned requirements has not yet been developed. Most evaluation procedures either require a degree of human involvement in order to be somewhat correlated to human judgement, or they require significant engineering effort. The evaluation methods, which we cover in this survey, can be categorized as follows: model the human judges, model the user behaviour, or use fine-grained methods, which evaluates a specific aspect of the dialogue system (e.g. its ability to adhere to a topic). Methods that model human judges rely on human judgements to be collected beforehand so as to fit a model which predicts the human rating. User behaviour models involve a significant engineering step in order to build a model which emulates the human behaviour. The finer-grained methods also need a certain degree of engineering, which depends on the feature being evaluated. The common trait of the evaluation methods covered in this survey is that they are coupled to the characteristics of the dialogue system that are being considered. That is, a task-oriented dialogue system is evaluated differently to a conversational dialogue system. 
%All the evaluation methods have in common that they depend on the characteristics of the dialogue system under consideration. 
\subsection{Modular Structure of this Article}
Different evaluation procedures have been proposed based on the characteristics of the dialogue system class. For instance, the evaluation of task-oriented systems exploits the highly structured dialogues. The goal can be precisely defined and measured to compute the task-success rate. On the other hand, conversational agents generate dialogues that are more unstructured, which can be evaluated on the basis of appropriateness of the responses; this has been shown to be difficult to automate. 
We introduce each type of dialogue system to highlight the respective characteristics and methods used to implement the dialogue system. With this knowledge, we introduce the most important concepts and methods developed to evaluate the respective class of dialogue system. In the following survey, we discuss each of the three classes of dialogue systems separately. Thus, Section \ref{sec:eval_task}:  \emph{Task Oriented Dialogue Systems}, Section \ref{sec:eval_non_task}: \emph{Conversational Agents}, and Section \ref{sec:eva_qa_dialogue}: \emph{Interactive Question Answering} can be read independently from each other.

\section{Task Oriented Dialogue System}
\label{sec:eval_task}
% !TEX root = survey_dialogue_evaluation.tex

\subsection{Characteristics}
\begin{figure}[h!]
  \begin{center}
    \includegraphics[width=0.6\textwidth]{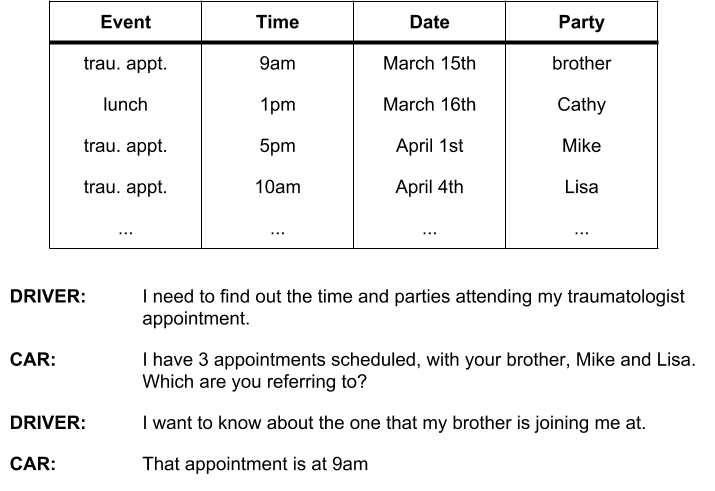}
    \caption{\label{fig:sample_dialogue}Example dialogue where the driver can query the agenda via a voice command \citep{mihail2017kv}. The dialogue system guides the driver through the various options.}
  \end{center}
\end{figure}

As the name suggests, a task-oriented dialogue system is developed to perform a clearly defined task. These dialogue systems are usually characterized by a clearly defined and measurable goal, a structured dialogue behaviour, a closed domain to work on and a focus on efficiency. Usually, the task involves finding information within a database and returning it to the user, performing an action, or retrieving information from its users. For instance, a restaurant information dialogue system helps the user to find a restaurant which satisfies the user's constraints. Furthermore, task-oriented dialogue systems also serve as interfaces to program APIs, which is often used in the Smart Home setting \citep{moller-etal-2004-inspire}. For example, an in-car entertainment dialogue system can be ordered to start playing music via voice commands or querying the agenda (see Figure \ref{fig:sample_dialogue} for an example).

The commonality is that the dialogue system infers the task constraints through the dialogue and retrieves the information requested by the user. For a ticket reservation system, the dialogue system needs to know the origin station, the destination, and the departure date and time. In most cases, the dialogue system is designed for a specific domain, such as restaurant information. The nature of these dialogue systems makes the dialogues both very structured and tailored. The ideal dialogue satisfies the user goal with as few interactions as possible. The dialogues are characterized by mixed initiatives, the user states its goal but the dialogue system proactively asks questions to retrieve the required constraints. 

\subsection{Dialogue Structure}
The dialogue structure for task-oriented systems is defined by two aspects: the content of the conversation and the strategy used within the conversation.
\paragraph{Content.} The content of the conversation is derived from the domain ontology. The domain ontology is usually defined as a list of slot-value pairs. For instance, Table \ref{tbl:domain-ontology} shows the domain ontology for the restaurant domain~\citep{novikova2017e2e}. Each slot has a type and a list of values, which the slot can be filled with. 
\begin{table}[h!]
\centering
\resizebox{0.7\textwidth}{!}{
\begin{tabular}[t]{@{}l|l|l}
\thead{Slot} & \thead{Type} & \thead{Example Values}\\
\hline
\hline
name & verbatim string & Alimentum, ..\\
eatType & dictionary & restaurant, pub, coffee shop\\
familyFriendly & boolean & yes, no\\
food & dictionary & Italian, French, English, ...\\
near & verbatim string & Burger King\\
area & dictionary & riverside, city center\\
customerRating & dictionary & \makecell[l]{1 of 5, 3 of 5, 5 of 5,\\ low, average, high}\\
priceRange & dictionary  & \makecell[l]{ \textless \pounds 20, \pounds 20-25, \textgreater \pounds 30 \\cheap, moderate, high}\\
\end{tabular}}
\caption{ Domain ontology of the E2E dataset \protect\citep{novikova2017e2e}. There are eight different slots (or attributes), each has a type and a set of values it can take.}
\label{tbl:domain-ontology}
\end{table}
 
\paragraph{Strategy.} While the domain ontology defines the content of the dialogue, the strategy to fill the required slots during the conversation is modelled as a sequence of actions \citep{austin1962how}. These actions are so-called \emph{dialogue acts}. A dialogue act is defined by its type (e.g. inform, query, confirm, and housekeeping) and by the list of arguments it can take. Each utterance corresponds to an action performed by an interlocutor. 

Table \ref{tbl:dialogue-acts} shows the dialogue acts proposed by \cite{Young:2010:HIS:1621140.1621240}.
\begin{table}[h!]
\begin{center}
\begin{tabular}{ll}
Dialogue Act                  & Description \\ \hline
hello$(a=x, b=y,..)$   & Open a dialogue and give info $a=x, b=y, ..$   \\
inform$(a=x, b=y,..)$  & Give information $a=x, b=y, ..$ \\
request$(a, b=x, ..)$  & Request value for $a$ given $b=x, ...$  \\
reqalts$(a=x,..)$      & Request alternative with $a=x,..$  \\
confirm$(a=x, b=y,..)$ & Explicitly confirm $a=x, b=y, ..$\\
confreq$(a=x, ..., d)$ & Implicitly confirm $a=x,..$ and request value of $d$ \\
select$(a=x, a=y)$     & Select either $a=x$ or $a=y$ \\
affirm$(a=x, b=y)$     & Affirm and give further info $a=x, b=y, ..$  \\
negate$(a=x)$           & Negate and give corrected value $a=x$  \\
deny$(a=x)$            & Deny that $a=x$  \\
bye$()$              &   Close a dialogue         
\end{tabular}
\caption{Dialogue acts proposed by \protect\cite{Young:2010:HIS:1621140.1621240}.}
\label{tbl:dialogue-acts}
\end{center}
\end{table}

For instance, the \emph{inform} act is used to inform the user about its arguments, i.e. inform(food = ``French'', area = ``riverside'') informs the user that there is a French restaurant at the riverside area. On the other hand, the \emph{request} act is used to request a value for a given list of slot-value pairs.

\begin{table}[h!]
\begin{center}
\resizebox{0.95\textwidth}{!}{
\begin{tabular}{ll}
Utterance & Dialogue Act \\ \hline
U: Hi, I am looking for somewhere to eat   & hello(task = find,type=restaurant)   \\
S: You are looking for a restaurant. What type of food?  & confreq(type = restaurant,food)\\
U: I’d like an Italian somewhere near the museum.  &  inform(food = Italian,near=museum) \\
S: Roma is a nice Italian restaurant near the museum. & inform(name = ‘‘Roma”, type = restaurant, food = Italian, near = museum) \\
U: Is it reasonably priced? & confirm(pricerange = moderate)\\
S: Yes, Roma is in the moderate price range. & affirm(name = ‘‘Roma”, pricerange = moderate) \\
U: What is the phone number?  & request(phone) \\
S: The number of Roma is 385456.    & inform(name = ‘‘Roma”, phone = ‘‘385456”)  \\
U: Ok, thank you goodbye.           & bye()  \\      
\end{tabular}}
\caption{Sample Dialogue and corresponding dialogue acts.}
\label{tbl:dialogue-acts-example}
\end{center}
\end{table}

Table \ref{tbl:dialogue-acts-example} shows an example dialogue with the corresponding dialogue acts. Each user utterance is translated into a dialogue act, and each dialogue act of the dialogue system is translated into an utterance in natural language. For instance, the utterance ‘‘Hi, I am looking for somewhere to eat” corresponds to the act of ‘‘hello”. The parameters describe the task that the user intends to solve, i.e. find a restaurant.
%The user starts by stating the task, namely, finding a restaurant. The system implicitly confirms the request and asks for the preferred food. Here, the dialogue system is asking questions to fill the food-type slot. After the user has provided the system with a few values for different slots, the system proposes a restaurant. Upon the proposal, the user asks to confirm another slot, namely, the price range. Then the user requests the phone number and the interaction ends. 
For a formal description of dialogue acts, refer to \cite{Traum1999,young2007cued}.

\subsection{Technologies}
We have just seen that content and strategy are the two main aspects driving the structure of a dialogue, but their influence reaches down to the different functionalities making a classic dialogue system architecture. It is composed of several parts which are built around the idea of modelling the dialogue as a sequence of actions. 

The central component is the so-called \emph{dialogue manager}. It defines the dialogue policy, which consists in deciding which action to take at each dialogue turn. The input to the dialogue manager is the current state of the conversation. The output of the dialogue manager is a dialogue act, which represents the system's action. Other components convert the user's input into a dialogue act and the dialogue manager's output into a natural language utterance. 

Usually, the user's input is processed by a natural language understanding (NLU) unit, which extracts the slots and their values from the utterance and identifies corresponding the dialogue act. This information is passed to the dialogue state tracker (DST), which infers the current state of the dialogue. Finally the output of the dialogue manager is passed to a natural language generation (NLG) component. 

Traditionally, these components were assembled into a pipelined architecture, but recent approaches based on trainable end-to-end neural networks offer a promising alternative. In the following, we briefly introduce the modules of the pipelined architecture and the deep neural network based approach.

\subsubsection{Pipelined Systems}
\label{subsec:slot}
\begin{figure}
  \begin{center}
    \includegraphics[width=0.8\textwidth]{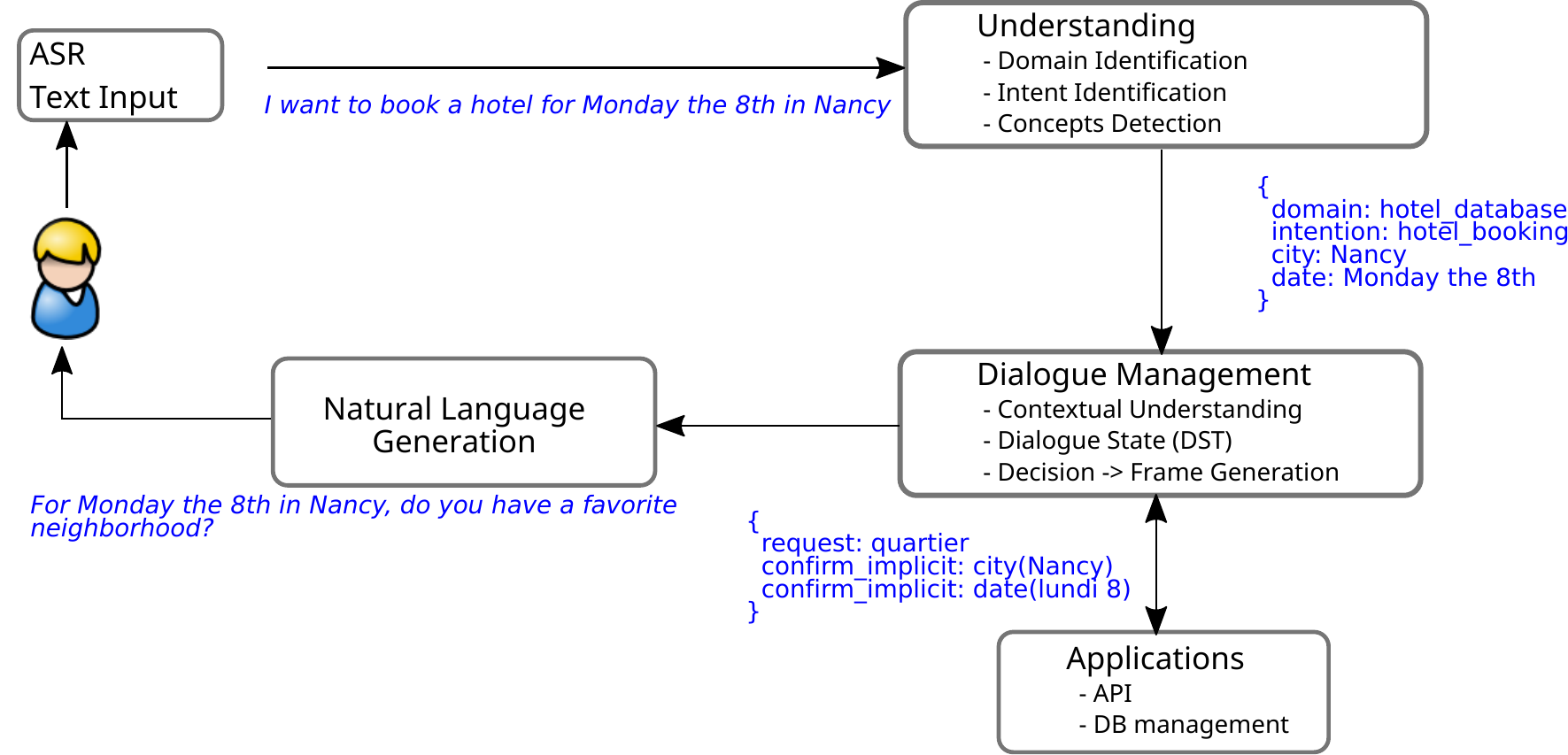}
    \caption{\label{fig:general-task-oriented}General overview of a task-oriented dialogue system.}
  \end{center}
\end{figure}

Usually, these four components are put into a pipelined architecture, where the output of one component is fed as the input into the next component (see Figure \ref{fig:general-task-oriented}). The input of the dialogue system is either a chat-interface or an automatic speech recognition (ASR) system. The input to the NLU unit is the utterance of the user in text format or, in the case of automatic speech recognition (ASR) a list of the N-best last user utterance transcriptions.
\paragraph{Natural Language Understanding.} The goal of the natural language understanding (NLU) unit is to detect the slot-value pairs expressed in the current user utterance. Since the early 2000s, the natural language understanding task is often seen as a set of subtasks~\citep{tur2011} as follows: (i) identification of domain (if multiple domains), (ii) identification of intents (that is, the question type, the dialogue act, etc.) and (iii) identification of the slots or concept detection. 

In an utterance such as, ``I want to book a hotel room for Monday, 8th", the domain is \emph{hotel}, the intent \emph{hotel booking} and the slot-value pair is \emph{date(Monday, 8th)}. The first two tasks are formalized as a classification task and any classification methods may be used. For concept detection one makes use of sequence labelling methods such as \emph{Conditional Random Field} (CRF)~\citep{Hahn.etAL:SLUJournal:2010} or recurrent neural network, typically bi-LSTM with CRF layer~\citep{yao2014spoken,Mesnil:2015:URN:2876369.2876380}. Recent methods propose to jointly learn the tasks of intent identification and concept detection~\citep{guo2014joint,zhang2016joint}. 
Usually, NLU is performed on classifying the intents that lie within the domain for which the dialogue system is developed for. \cite{larson-etal-2019-evaluation} introduce an out-of-scope intent classification task, where the NLU system is trained to detect if a user intent does not lie within the scope of the dialogue systems' capabilities. 

\begin{figure}
  \begin{center}
    \includegraphics[width=0.75\textwidth]{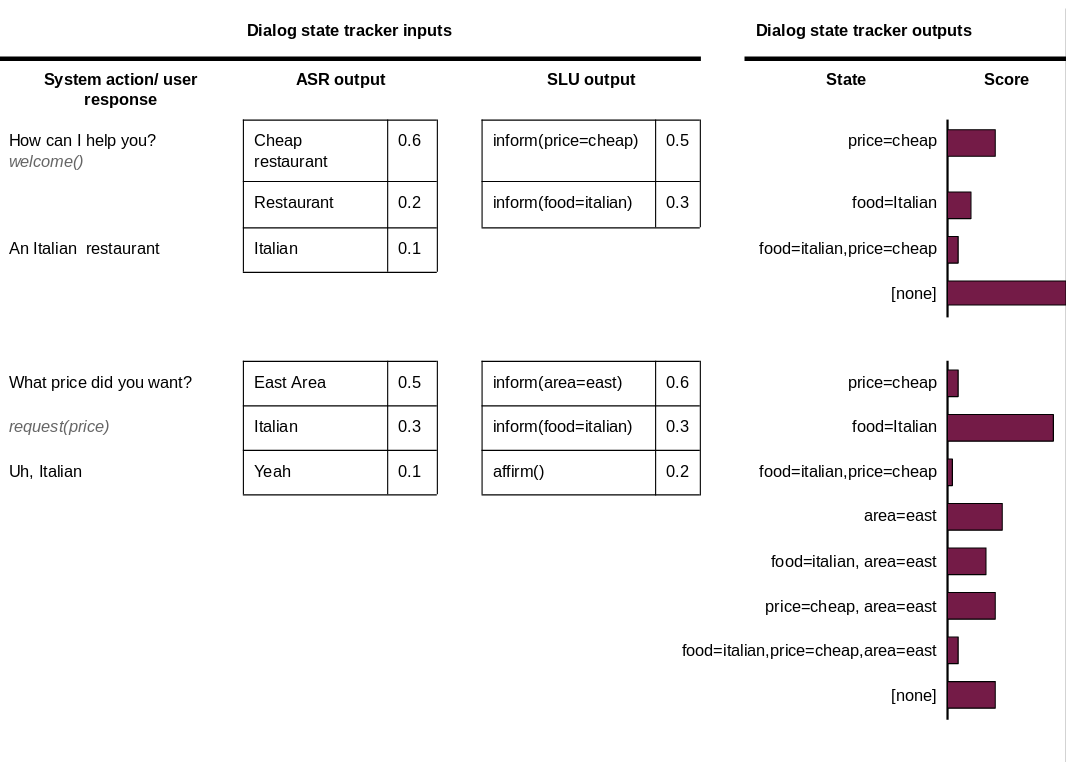}
    \caption{\label{fig:dstc}Overview of a DST module. The input to the DST module is the combined output of the ASR and the NLU model.}
  \end{center}
\end{figure}

\paragraph{Dialogue State Tracking.} The Dialogue State Tracker (DST) infers the current \emph{belief state} of the conversation, given the dialogue history up to the current point $t$ \citep{the-dialog-state-tracking-challenge-series-a-review}. The current belief state encodes the user's goal (e.g. which price range the user prefers) and the relevant dialogue history, i.e. it is an internal representation of the state of the conversation. It is important to take the previous belief states into account in order to handle misunderstandings. For instance, in Figure \ref{fig:dstc}, the confidence that the user wants an Italian restaurant is low. In the successive turn, the ASR system still gives low confidence to the Italian restaurant. However, since the state tracker takes into account that the Italian restaurant could have been mentioned in the previous turn, it assigns a higher overall probability to it. 

The main challenge for the DST module is to handle the uncertainty, which stems from the errors made by the ASR module and the NLU unit. Typically, the output of the DST unit is represented as a probability distribution over multiple possible dialogue states $b(s)$, which provides a representation of the uncertainty. Generative methods have been widely used to manage this task, for example, dynamic Bayesian network (DBN) along with a beam search~\citep{young2007hidden}. Those methods present some limits which are widely discussed in~\cite{metallinou-bohus-williams:2013:ACL2013}, the most important being that all the correlations in the input features have to be modeled (even the unseen cases). 

Discriminative models were then proposed to overcome these limits. \cite{metallinou-bohus-williams:2013:ACL2013} proposed to use a linear classifier with the dialogue history present in the input features. Whereas \cite{henderson2013deep} proposed to map directly the ASR hypotheses onto a dialogue state by means of recurrent neural networks. This way, both NLU and DST were integrated into a single function. Nowadays, neural approaches are becoming more and more popular \citep{mrk2017dst}.

\paragraph{Strategy.} The strategy is learned by the dialogue manager. The input is the current belief state $b(s)$ computed by the DST module. The DM generates the next action of the system, which is represented as a dialogue act. In other words, based on the current turn values and on the value history the system performs an action (e.g. retrieve data from a database, ask for a missing information, etc.). Deciding which action to take is part of the dialogue control. 

In earlier systems, the dialogue control was based on a finite-state automaton in which the nodes represent the questions of the system and the transitions the possible user's answers. This method, while being rigid, is efficient when the domain and the task are simple. It has been widely used to design dialogue systems and many toolkits are available such as the one from the Center for Spoken Language Understanding~\citep{cole1999tools} or VoiceXML.\footnote{See \url{https://www.w3.org/TR/voicexml20/}} The main issue is the rigid dialogue structure as well as the tendency to be error-prone. In fact, such a system does not model discourse phenomena like ellipsis (a part of the sentence structure that can be inferred from the context is omitted) or anaphoric references (which can be resolved only in a given context). 

To overcome these inefficiencies, a dialogue manager is designed to keep track of the interaction history and controls the dialogue strategy. This is called frame-based dialogue control and management. Frame-based techniques rely on schemas specifying what the system has to solve instead of representing what the system has to do and when. This allows for dialogue to be more flexible and the possibility to handle errors~\citep{mctear2005handling,ritel-interspeech07}. 

Initially, dialogue managers were implemented using rule-based approaches. When data had become available in sufficient amount, data-driven methods were proposed for learning dialogue strategies from data. The dialogue is represented as a Markov decision problem (MDPs), following the intuition that a dialogue can be represented as a sequence of actions \citep{levin1998mdp,NIPS1999_1775}. These actions are referred to as \emph{speech acts} or \emph{dialogue acts}~\citep{austin1962how,searle1969speech,searle1975indirect}. However, MDPs cannot handle uncertainty coming from speech recognition errors \citep{young2013pomdp_overview}. 

Thus, partially observable MDPs (POMDP) were adopted, as they introduce the belief state, which models the uncertainty of the current state \citep{peak2006rl_ds,Lemon:2012:DMA:2412075,young2013pomdp_overview}. Although this alleviated the problem of hand-crafting the dialogue policy, the domain ontology still needs to be manually created. Furthermore, these dialogue systems are trained on a static and well-defined domain, once trained the policy works only on this domain. Finally, the dialogue systems need large amounts of data to be trained efficiently, mostly using user simulation for training \citep{schatzmann:2006:us_survey}. Beyond user simulations, \cite{gasic2011gp_pomdp} showed that online policy learning based on crowdsourcing is a valid alternative. 

To mitigate the issues arising from the lack of data, \cite{gasic2011gp_pomdp} applied Gaussian processes for POMDP-based optimization \citep{Engel:2005:RLG:1102351.1102377}, which exploits the correlation between different belief states and speeds up the learning process. The authors showed that a reasonable policy can be learned with online user feedback after a few hundred dialogues. \cite{gasic2013domain,gasic2014domain} showed that it is possible to adapt the policy if the domain is extended dynamically. Note also the work of \cite{wang2015domain} which aims at enabling domain-transfer by introducing a domain-independent ontology parametrisation framework.
 
\paragraph{Natural Language Generation.} The natural language generation (NLG) module translates the dialogue act represented in a semantic frame into an utterance in natural language~\citep{bangalore2001natural}. The task of NLG is usually divided into separate subtasks such as content selection, sentence planning, and surface realization \citep{Stent:2004:TSP:1218955.1218966}. Traditionally, the task has been solved by relying on rule-based methods and canned texts. %, but these methods do not scale and are hardly adaptable to new domains.
Statistical methods were also proposed and used, such as phrase-based NLG with statistical language models~\citep{mairesse2010phrase} or CRF based on semantic trees~\citep{dethlefs2013conditional}. Recently, deep learning techniques have become more prominent for NLG. With these techniques, there now exists a large variety of different network architectures, each addressing a different aspect of NLG; \cite{wensclstm15} propose an extension to the vanilla LSTM \citep{hochreiter1997} to control the semantic properties of an utterance, whereas \cite{hu2017} use variational autoencoder (VAE) and generative adversarial networks to control the generation of texts by manipulating the latent space; \cite{mei2016talk} employ an encoder-decoder architecture extended by a coarse-to-fine aligner to solve the problem of content selection; \cite{N16-1015} apply data counter-fitting to generate out-of-domain training data for pretraining a model where there is little in-domain data available; \cite{semeniuta2017hybrid} and \cite{bowman2015generating} use a VAE trained in an unsupervised fashion on large amounts of data to sample texts from the latent space; and \cite{P16-2008} use a sequence-to-sequence model with attention to generate natural language strings as well as deep syntax dependency trees from dialogue acts.

\subsubsection{End-to-end trainable Systems}
\label{subsec:ent-to-end}
%Discussion about e2e systems with task-oriented, will be rather short as it is very new
Traditionally, task-oriented dialogue systems were designed along the pipelined architecture, where each module has to be designed, trained, and evaluated separately. There are several drawbacks to this approach. As the architecture is modular, each component needs to be designed separately, which involves lots of hand-crafting, the costly generation of annotated data for each module, and training each component \citep{wen2017e2e_dialog}. Furthermore, the pipelined architecture leads to the propagation and amplification of errors through the pipeline as each module depends on the output of the previous module \citep{li2017e2e_dial,liu2018e2eds}. 

Related to the architecture there is a credit assignment problem, as the dialogue system is evaluated as a whole, it is hard to determine what module is responsible for which reward. Furthermore, this architecture leads to interdependence among the modules, i.e. when one module is changed, all the subsequent modules need to be adapted as well \citep{zhao2016e2edm}. 

Finally, the slot-filling architecture, which is often used, makes these systems inherently hard to scale to new domains since there is a need to hand-craft the representation of the state and action space \citep{DBLP:journals/corr/BordesW16}. 

To overcome these limitations, current research focuses on end-to-end trainable architectures where the dialogue system is trained as a single module. \cite{wen2017e2e_dialog} %the authors
model the dialogue as a sequence to sequence mapping, where the traditional pipeline elements are modelled as interacting neural networks. The policy network takes as input the results form the intent network, the belief tracker network, the database operator and selects the next action, based on the selected action, the generation network produces the output utterance.

 \cite{DBLP:journals/corr/BordesW16} propose a set of synthetic tasks to evaluate the feasibility of end-to-end models in the task-oriented setting, for which they use a memory network to model the conversation. These approaches learn the dialogue policy in a supervised fashion from the data. In contrast the work by \cite{li2017e2e_dial,zhao2016e2edm} train the system using reinforcement-learning. Note that all these approaches rely on huge amounts of training data.

\subsection{Evaluation}
The evaluation of task-oriented dialogue systems is built around the structured nature of the interaction. Two main aspects are evaluated, which have been shown to define the quality of the dialogue: task-success and dialogue efficiency. Two main metrics of evaluation methods have been proposed:
\begin{itemize}
\item User Satisfaction Modelling: Here, the assumption is that the usability of the system can be approximated by the satisfaction of its users, which can be measured by questionnaires. These approaches aim to model the human judgements, i.e. creating models which give the same ratings as the human judges. First, a human evaluation is performed where subjects interact with the dialogue system. Afterwards, the dialogue system is rated via questionnaires. Finally, the ratings are used as target labels to fit a model based on objectively measurable features (e.g. task success rate, word error rate of the ASR system).
\item User Simulation: Here, the idea is to simulate the behaviour of the users. There are two applications of user simulation: firstly, to evaluate a functioning system with the goal of finding weaknesses and secondly, the user simulation is used as an environment to train a reinforcement-learning based system. The evaluation in the latter is based on the reward achieved by the dialogue manager under the user simulation. 
\end{itemize}
Both these approaches rely on measuring task-success rate and dialogue efficiency. Before we introduce the methods themselves, we will go over the ways to measure performance along these two dimensions. 

\begin{figure}
  \begin{center}
    \includegraphics[width=0.75\textwidth]{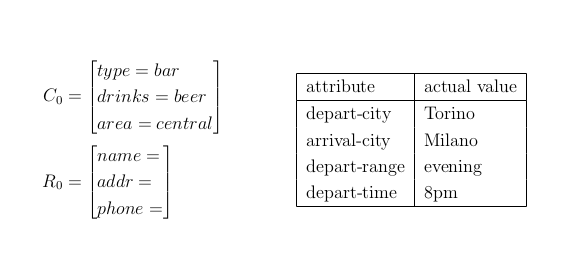}
    \caption{\label{fig:goal_constratins_req} Examples of goals from \protect\cite{Schatzmann:2007:AUS:1614108.1614146}  and \protect\cite{Walker:1997:PFE:979617.979652}. Where $C_0$ denotes the information constraints, i.e. which information is to be retrieved (a bar that serves beer in the city center). $R_0$ denotes the set of requests, i.e. the information the user wants (name, address, and phone number).}
  \end{center}
\end{figure}

\paragraph{Task-Success Rate.} The goal or the task of the dialogue can be split into two parts \citep{Schatzmann:2007:AUS:1614108.1614146} (see Figure \ref{fig:goal_constratins_req}) as follows: 
\begin{itemize}
\item Set of Constraints, which define the target information to be retrieved. For instance, the specifications of the venue (e.g. a bar in the central area, which serves beer) or the travel route (e.g. ticket from Torino to Milano at 8pm). 
\item Set of Requests, which define what information the user wants. For instance the name, address and the phone number of the venue. 
\end{itemize}
The task-success rate measures how well the dialogue system fulfills the information requirements dictated by the user's goals. For instance, this includes whether the correct type of venue has been found by the dialogue system and whether the dialogue system returned all the requested information. One possibility to measure this is via a confusion matrix (see Table \ref{tab:confusion_matrix}), which represents the errors made over several dialogues. Based on this representation, the Kappa coefficient \citep{carletta:1996:CL1996v22n2} can be applied to measure the success (see \cite{powers:2012:EACL2012} for Kappa shortcomings). 

\begin{table}[h!]
  \begin{center}
    \resizebox{0.95\textwidth}{!}{%
      \begin{tabular}{|c|cccc|cccc|cccr|cccc|}
        \hline %
        \multicolumn{17}{|c|}{KEY} \\ \hline
             & \multicolumn{4}{c|}{DEPART-CITY}                        & \multicolumn{4}{c|}{ARRIVAL-CITY}                      & \multicolumn{4}{c|}{DEPART-RANGE}  & \multicolumn{4}{c|}{DEPART-TIME}                       \\ \hline
        DATA & v1           & v2          & v3          & v4          & v5          & v6          & v7          & v8          & v9          &   &   & v10         & v11         & v12         & v13         & v14         \\ \hline
        v1   & \textbf{22}  &             &  1          &             &  3          &             &             &             &             &   &   &             &             &             &             &             \\
        v2   &              & \textbf{29} &             &             &             &             &             &             &             &   &   &             &             &             &             &             \\
        v3   &  4           &             & \textbf{16} &  4          &             &             &  1          &             &             &   &   &             &             &             &             &             \\
        v4   &  1           &  1          &  5          & \textbf{11} &             &             &  1          &             &             &   &   &             &             &             &             &             \\ \hline
        v5   &              &             &             &             & \textbf{20} &             &             &             &             &   &   &             &             &             &             &             \\
        v6   &              &             &             &             &             & \textbf{22} &             &             &             &   &   &             &             &             &             &             \\
        v7   &              &             &             &             &  1          &  1          & \textbf{20} &  5          &             &   &   &             &             &             &             &             \\
        v8   &              &             &             &             &  1          &  2          &  8          & \textbf{15} &             &   &   &             &             &             &             &             \\ \hline
        v9   &              &             &             &             &             &             &             &             & \textbf{45} &   &   & 10          &             &             &             &             \\
        v10  &              &             &             &             &             &             &             &             &  5          &   &   & \textbf{40} &             &             &             &             \\ \hline
        v11  &              &             &             &             &             &             &             &             &             &   &   &             & \textbf{20} &             &  2          &             \\
        v12  &              &             &             &             &             &             &             &             &             &   &   &             &  1          & \textbf{19} &  2          &  4          \\
        v13  &              &             &             &             &             &             &             &             &             &   &   &             &  2          &             & \textbf{18} &             \\
        v14  &              &             &             &             &             &             &             &             &             &   &   &             &  2          &  6          &  3          & \textbf{21} \\ \hline
        sum  & 30           & 30          & 25          & 15          & 25          & 25          & 30          & 20          & 50          &   &   & 50          & 25          & 25          & 25          & 25          \\ \hline
      \end{tabular}
    }
    \caption{\label{tab:confusion_matrix} Confusion matrix from \protect\cite{Walker:1997:PFE:979617.979652}. For each key (e.g. depart-city) a confusion matrix is created, which denotes the expected values (row) and the values produced by the dialogue system  (columns). For instance, if it was expected that the dialogue system returns the train schedule from Torino to Milano but it confused the depart-city with Verona, then this is counted as an error. }
  \end{center}
\end{table}

\paragraph{Dialogue Efficiency.} Dialogue efficiency or dialogue costs are measures which are related to the length of the dialogue~\citep{Walker:1997:PFE:979617.979652} . For instance, the number of turns or the elapsed time are such measures. More intricate measures could include the number of inappropriate repair utterances or the number of turns required for a sub-dialogue to fill a single slot.  

In the following, we introduce the most important research for both of the aforementioned evaluation procedures. Finally, we briefly cover the evaluation methods employed on the subsystems of the pipleline. However, the main focus of this review is the evaluation of the dialogue system's behaviour. 

% !TEX root = survey_dialogue_evaluation.tex
\subsubsection{User Satisfaction Modelling}
\label{sce:user_satisfaction_modelling}
User satisfaction modelling is based on the idea that the usability of a system can be approximated by the satisfaction of its users. The research in this area is concerned with three goals: measure the impact of the properties of the dialogue system on the user satisfaction (explainability requirement), automate the evaluation process based on these properties (automation requirement), and use the models to evaluate different dialogue strategies (differentiability requirement). Usually, a predictive model is fit, which takes the properties as input and uses the human judgements as target variable. Thus, modelling the user satisfaction as either a regression or a classification task.
There are different approaches to measure the user satisfaction, which are based on two questions: who evaluates the dialogue and at which granularity is the dialogue evaluated? The first question allows for two groups; either the dialogue is evaluated by the users themselves or by objective judges. The second question allows for different points on a spectrum. On one end, the evaluation takes place on the dialogue level, on the other end the evaluation takes place at the exchange level.
The question of who evaluates the dialogue is often especially at the centre of discussion. Here, we will shortly summarize the main points.

\paragraph{User or Expert ratings.}
There are three main criticisms regarding the judgments made by users: 
\begin{itemize}
\item Reliability: \cite{evanini2008:caller_experience} state as a main argument that users tend to interpret the questions on the questionnaires differently, thus making the evaluation unreliable. \cite{gasic2011gp_pomdp} noted that also in the lab setting, where users are given a predefined goal, users tend to forget the task requirements, thus, incorrectly assessing the task success. Furthermore, in the in-field setting, where the feedback is given optionally, the judgements are likely to be skewed towards the positive interactions.
\item Cognitive demand: \cite{SCHMITT201512} note that rating the dialogue puts more cognitive demand on users. This is especially true if the evaluation has to be done at the exchange level. This would falsify the judgments about the interaction.
\item Impracticability: \cite{ultes2013quality} note the impracticability of having a user rate the live dialogue, as he would have to press a button on the phone, or have a special installation to give feedback. 
\end{itemize}
\cite{ultes2013quality} analyzed the relation between the user ratings and ratings given by objective judges (called \emph{experts}). Especially, they investigated if the ratings from the experts could be used to predict the ratings of the users. Their results showed that the user ratings and the expert ratings are highly correlated with a Spearman's $\rho$ score of $\rho=0.66 (p < 0.01) $. Thus, expert ratings can be used as replacement for user judgments. Furthermore, they trained classifiers using the expert rating as targets and evaluated on the user ratings as targets. The best performing classifier achieved an \label{UARdef} unweighed average recall (UAR) of 0.34 compared to the best classifier trained on user satisfaction, which achieved $UAR=0.5$. These results indicate that it is not possible to precisely predict the user satisfaction. However the correlation scores show that the predicted scores of both models correlate equally to the user satisfaction $p = 0.6$. Although the models cannot be used to exactly predict the user satisfaction, the authors showed that the expert ratings are strongly related to user ratings.\\

In the following, we present different approaches to user satisfaction modelling. We cover the most important research for each of the various categories.

\paragraph{PARADISE Framework.}
\label{subsec:paradise}
PARADISE (PARAdigm for DIalog System Evaluation) \citep{Walker:1997:PFE:979617.979652} is the most known evaluation framework proposed for task-oriented systems. It is a general framework, which can be applied to any task-oriented system, since it is domain-independent. It belongs to the evaluation methods which are based on user ratings on the dialogue level, although it allows for evaluations of sub-dialogues. 

Originally, the motivation was to produce an evaluation procedure, which can distinguish between different dialogue strategies.  At that time, the most widely used automatic approach was based on the comparison of utterances with a reference answer \citep{hirschman1990beyond}. Methods based on comparisons to reference answers suffer from various drawbacks: they cannot discriminate between different strategies, they are not capable of attributing the performance on system specific properties, and the approach is not generalizable to other tasks.

The main idea of PARADISE is to  combine different measures of performance into a single metric, and in turn assess the contribution of each of these measures to the final user satisfaction. PARADISE originally uses two objective measures for performance: task-success and measures that define the dialogue cost (as explained above). 

\begin{figure}[h!]
  \begin{center}
    \includegraphics[width=0.9\textwidth]{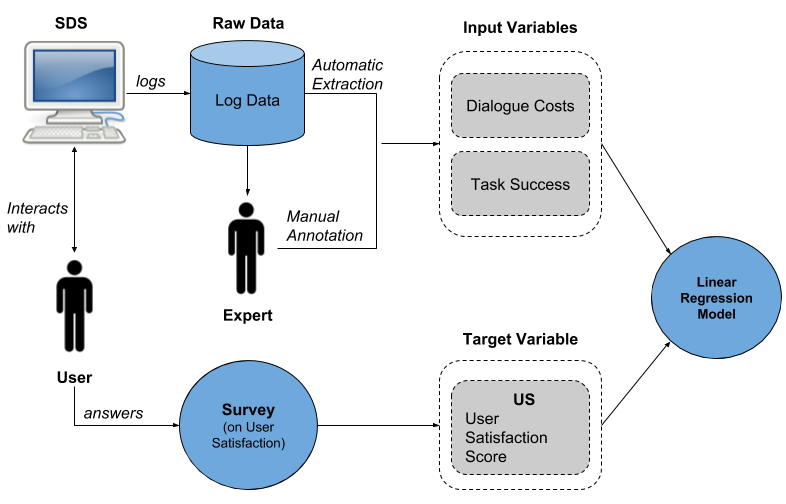}
    \caption{\label{fig:paradise_overview} PARADISE Overview \citep{SCHMITT201512}}
  \end{center}
\end{figure}

An overview of the PARADISE framework is depicted in Figure \ref{fig:paradise_overview}. The user interacts with the dialogue system and completes a questionnaire after the dialogue ends. From the questionnaire, a user satisfaction score is computed, which is used as the target variable. The input variables to the linear regression models are extracted from the logged conversation data. The extraction can be done automatically (e.g. for task-success as discussed above) or manually by an expert (e.g. for inappropriate repair utterances). Finally, a linear regression model is fitted to predict the user satisfaction for a given set of input variables.  
%\begin{itemize}
%\item Task-success is measured by how well the agent and the user achieve the required information requirements. This requires two steps: define a representation of the task and defining a measurement. The task is represented via a so-called attribute-value matrix (AVM). The AVM is closely related to the frames presented above, i.e. it contains a list of attributes and possible values, which have to be exchanged during the dialogue. For example, for a timetable information system, it should know the departure-and arrival city as well as the time of departure. A goal is then defined as retrieving the timetable for the train departing from Torino to Milan at 8pm. The task-success is measured by how well the interaction achieves the requirements of the goal. For this, the Kappa coefficient is computed. 
%\item Dialogue Costs: these measures are related to the behaviour of the bot. Among others, they include: numbers of turns to complete the task, number of inappropriate utterances, number of repair utterances. PARADISE also allows to calculate costs at the sub-dialogue level, where each sub-dialogue is the dialogue used to achieve a subtask - e.g. getting the departure city correctly. The AVM can be used to measure the number of utterances used to achieve a certain subtask.  
%\end{itemize}

Thus, PARADISE models the (subjective) performance of the system with a linear combination of objective measures (task-success and dialogue costs). Applying multiple linear regressions showed that only the task-success measure and the number of repetitions are significant. 
In a follow-up study \citep{walker_kamm_litman_2000}, the authors further investigated PARADISE's ability to generalize to other systems and user populations and its predictive power. For this, they applied PARADISE on three different dialogue systems: ELVIS (a dialogue system for accessing emails), ANNIE (a dialogue system for voice dialing and messaging), and TOOT (a dialogue system for accessing train schedules). In a large-scale user study, they collected 544 dialogues over 42 hours of speech. For these experiments, the authors worked with an extended number of quality measures: e.g. number of barge-ins (i.e. sudden interruption by the user), number of cancel operations, number of help requests. A survey at the end of the dialogue was used to measure the user satisfaction. The survey asked about various aspects: e.g. speech recognition performance, ease of the task, if the user would use the system again. Based on the survey, the user satisfaction score is computed and used as the target variable to train the PARADISE framework as described above. Table \ref{tab:paradise_eval} shows the generalization scores of PARADISE for different scenarios. 

\begin{table}[h!]
  \begin{center}
    \resizebox{0.95\textwidth}{!}{%
      \begin{tabular}{rrrr}
        \hline \hline %
        Training Set & $R^2$ Training (SE) & Test Set & $R^2$ Test (SE) \\ \hline \hline %
        %ANNIE 90\%        & 0.50 (0.009)             & ANNIE 10\%    & 0.40 (0.07)          \\
        %ELVIS 90\%        & 0.39 (0.003)             & ELVIS 10\%    & 0.43 (0.03)          \\
        %TOOT 90\%         & 0.56 (0.014)             & TOOT 10\%     & 0.54 (0.05)          \\
        ALL 90\%          & 0.47 (0.004)             & ALL 10\%      & 0.50 (0.035)         \\
        ELVIS 90\%        & 0.42                     & TOOT          & 0.55                 \\
        ELVIS 90\%        & 0.42                     & ANNIE         & 0.36                 \\
        NOVICES           & 0.47                     & ANNIE EXPERTS & 0.04                 \\
        %MRS $\le$ 0.95    & 0.46                     & MRS $>$ 0.95  & 0.23                 \\
        \hline \hline %
      \end{tabular}
    }
    \caption{\label{tab:paradise_eval} Predictive power of PARADISE. Where ALL denotes that the collection of all the annotated data from the three different systems. The distinction between NOVICES and EXPERTS denotes the level to which the test subjects were instructed to use the dialogue system. }
  \end{center}
\end{table}

According to these scores, we obtain the following observations:
\begin{itemize}
\item A linear regression model is fitted on $90\%$ of the data and evaluated on the remaining $10\%$. The results show that the model is able to explain $R^2 = 50\%$ of the variance, which is considered to be a good predictor by the authors. 
%A linear regression model was trained on the data for one dialogue system and evaluated on the test set containing data for the same dialogue system. The results for this condition vary between $0.4$ and $0.54$, which is considered to be a good predictor by the authors. 
\item Training the regression model on the data for one system and evaluating the model on the data for another dialogue system (e.g. train on the ELVIS data and evaluate on the TOOT data) show high variability as well. The evaluation on the TOOT system data yields much higher scores than evaluating on the ANNIE data. These results show that the model is able to generalize to data of other dialogue systems to a certain degree. 
\item The evaluation of the generalizability of the model across different populations of users yields a negative result. When trained on dialogue data from conversation by novice users (NOVICES), the linear model is not capable of predicting the scores by experienced users (ANNIE EXPERTS) of the dialogue system.
\end{itemize}
The PARADISE framework is not only able to find the factors, which have the most impact on the rating, it is also capable of predicting the ratings. However, the experiments also revealed that the framework is not capable of distinguishing between different user groups. This result was confirmed by \cite{engelbrecht2008analysis}, which tested the predictive power of PARADISE for individual users.

\paragraph{User satisfaction at the exchange level.}
\label{iq_section}
In contrast to rating the dialogue as a whole, in some cases it is important to know the rating at each point in time. This is especially useful for online dialogue breakdown detection. There are two approaches to modelling the user satisfaction at the exchange level: annotate dialogues at the exchange level either by users \citep{Engelbrech2009modeling} or by experts \citep{higa2010issues,SCHMITT201512}. Different models can be fitted with the sequential data: Hidden Markov Models (HMM), Conditional Random Fields or Recurrent Neural Networks are the most obvious choice, but also SVM based approaches are possible.  

\cite{Engelbrech2009modeling} model user satisfaction as a continuous process evolving over time, where the current judgment depends on the current dialogue events and the previous judgments. Users interacted with the dialogue system and judged the dialogue after each turn on a 5-point scale using a number pad. An HMM was trained based on these target values and annotated dialogue features. Some input features were manually annotated, which is not a reasonable setting for online breakdown detection. 

\cite{higa2010issues} modelled the evaluation similarly as in \cite{Engelbrech2009modeling}. In their study, they evaluated different models (HMM and CRF), different measures to evaluate the trained model, and addressed the question of subjectivity of the annotators. The input features to the model were the dialogue acts and the target variables were the annotations by experts, which listened to the dialogue. The low inter-rater agreement and the fact of only using dialogue acts as inputs made the model perform only marginally better than the random baseline. 

A different approach was taken by \cite{hara2010estimation}, who relied on dialogue-level ratings, but trained the model on n-grams of dialogue-acts. More precisely, they used as input features $n$ consecutive dialogue acts and used the dialogue-level rating as target variable (on a 5-point scale and an extra class to denote unsuccessful task). The model achieved an accuracy of only $34.4\%$ using a 3-gram model. Further testing yielded that the model is able to predict the task-success with an accuracy of $94.7\%$.

These approaches suffer from the following problems: they either rely on manual feature extraction, which is not useful for online breakdown detection or they used only dialogue acts as input features, which does not cover the whole dialogue complexity. Furthermore, the approaches had issues with data annotation, either having low inter-rater agreement or using dialogue-level annotation. 
\cite{SCHMITT201512} addressed these issues by proposing Interaction Quality (see next paragraph) as approximation to user ratings at the exchange level.

\paragraph{Interaction Quality.}
Interaction Quality is a metric proposed by \cite{SCHMITT201512} with the goal to allow the automatic detection of problematic dialogue situations. The approach is based on letting experts rate the quality of the dialogue at each point in time - the median rating of several expert ratings at the exchange level is called Interaction Quality. The experiments in this study were conducted using the \emph{Let's Go} bus information system \cite{black2009challenge}. 

\begin{figure}[h!]
  \begin{center}
    \includegraphics[width=0.9\textwidth]{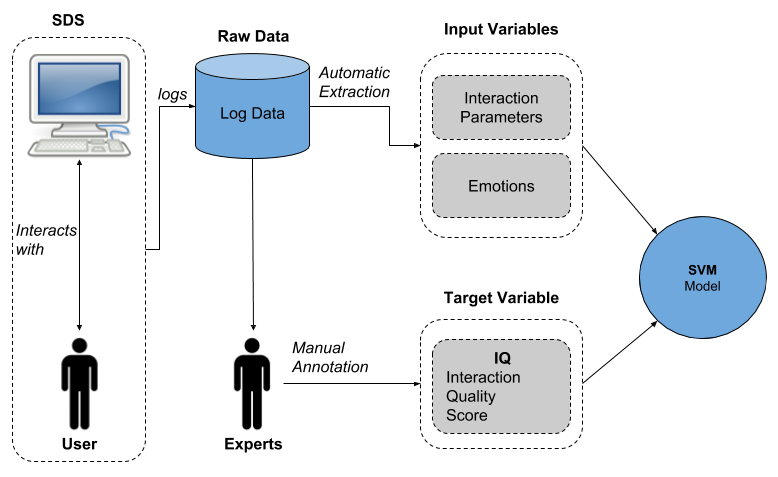}
    \caption{\label{fig:iq_overview} Overview of the Interaction Quality procedure \citep{SCHMITT201512}.}
  \end{center}
\end{figure}

Figure \ref{fig:iq_overview} shows the overview of the Interaction Quality procedure. The user interacts with the dialogue system and the conversation's relevant data is logged. From the logs, the input variables are automatically extracted. The target variables are manually annotated by experts, from which the target variable is derived. Based on the input and target variables, a support vector machine (SVM) is fitted. 

Interaction Quality is meant to approximate user satisfaction. In this study, the authors showed that Interaction Quality is an objective and valid approximation to user satisfaction, which is easier to obtain. This is especially important for in-field evaluations of dialogue systems, which are practically infeasible to be rated by users at the exchange level. Thus, it is important that in-field dialogues can be rated by experts at the exchange level. The challenge is to make sure that the ratings are objective, i.e. to eliminate the subjectivity of the experts as much as possible.

Since there is no possibility to gather user satisfaction scores at the exchange level from in-field conditions, the authors relied on user satisfaction scores from lab experiments and Interaction Quality scores over dialogues from both in-field and lab conditions. For the lab experiments, users interacted with the \emph{Let's Go} bus information system \citep{black2009challenge} and used a special device to rate the dialogue after each turn. These scores are referred to as user satisfaction. The dialogues were then rated by experts on the exchange level. These ratings are referred to as Interaction Quality. The authors found a strong correlation (Spearman's $\rho=0.66$) between Interaction Quality and user satisfaction in the lab environment, which means that Interaction Quality is a valid substitute for user satisfaction. In order to assess if Interaction Quality is a valid measure for rating in-field conversations, experts rated 200 dialogues from the \emph{Let's Go Field Corpus} \citep{schmitt2012lets_go} and measured the agreement among the experts. The experts achieved  a strong correlation (Spearman's $\rho=0.72$). 

Based on these Interaction Quality scores a predictive model is trained to automatically judge the dialogue at any point in time. In order to automatically predict Interaction Quality, the input variable need to be automatically extractable from the dialogue system. From each subsystem of a task-oriented dialogue system (Figure \ref{fig:general-task-oriented}), various values are extracted (AUTO features). Additionally, the authors experimented with hand-annotated features such as emotions (EMO) and user specific features (USER), such as age or gender, as well as semi-automatically annotated data such as the dialogue acts (similar to \cite{higa2010issues}). Based on these input variables, the authors trained various SVMs, one for each target variable, namely Interaction Quality for both in-field and the lab data as well as the user satisfaction label for the lab data. Table \ref{tab:iq_eval_small} shows the scores achieved for the various target variables and input feature groups. 

\begin{table}[h!]
  \begin{center}
    \resizebox{0.6\textwidth}{!}{%
    \begin{tabular}{@{}llll@{\hspace{25pt}}lll@{\hspace{25pt}}lll}
      \hline %
        Feature Set              & \emph{$IQ_{field}$} &  \emph{$IQ_{lab}$} &  \emph{$US_{lab}$} \\ \hline
        ASR               & 0.753  & 0.811  & 0.625 \\
        AUTO              & 0.776  & 0.856  & 0.668  \\
        AUTO + EMO        & 0.785   & 0.856 & 0.669\\
        AUTO + EMO + USER & -      & 0.888   & 0.741 \\ \hline
    \end{tabular}
  }
    \caption{\label{tab:iq_eval_small} Model performance (in terms of $\rho$) on the test set. \protect\cite{SCHMITT201512}. ASR denotes the features by the automatic speech recognition system. AUTO denotes automatically extracted features from the dialogue system pipeline (e.g. dialogue acts). EMO denotes features that capture the users emotions (e.g. anger). USER denotes user specific features (e.g. age, gender).}
  \end{center}
\end{table}

The in-field Interaction Quality model ($IQ_{field}$) achieves a correlation of $\rho = 0.776$ to the human judges, based on the automatically extracted features, with the ASR features alone the correlation score lies at $\rho = 0.753$. The addition of the emotional and user -specific features do not increase the scores significantly. A similar behaviour is measured for the lab Interaction Quality model ($IQ_{lab}$), which achieves high scores with ASR features alone ($\rho = 0.856$) and profits only marginally from the inclusion of the emotional features. However, the model improves when including user specific features ($\rho = 0.894$). The lab based user satisfaction model ($US_{lab}$) achieves lower scores with $\rho = 0.668$ for the automatic features. 

\begin{table}[h!]
  \begin{center}
    \begin{tabular}{llll}
      \hline %
      Feature set & Test & Train & $\rho$ \\ \hline
      Auto & \emph{$US_{lab}$}   & \emph{$IQ_{lab}$}   & 0.667 \\
      Auto & \emph{$IQ_{lab}$}   & \emph{$IQ_{field}$} & 0.647  \\
      Auto & \emph{$IQ_{field}$} & \emph{$IQ_{lab}$}   & 0.696 \\ \hline
    \end{tabular}
    \caption{\label{tab:iq_cross_eval} Model performance (in terms of $\rho$, $\kappa$ and UAR) on the test set. \protect\cite{SCHMITT201512}}
  \end{center}
\end{table}

Table \ref{tab:iq_cross_eval} shows the cross model evaluation. The $IQ_{field}$ model can be used to predict $IQ_{lab}$ labels and vice versa ($\rho \sim 0.66$). Furthermore, the $IQ_{lab}$ model is able to predict the $US_{lab}$ variable. These results show that Interaction Quality is a good substitute to user satisfaction and that the models based on Interaction Quality yield high predictive performance when trained on the automatically extracted features. This allows to evaluate an ongoing dialogue in real-time at the exchange level and ensures high correlation to the actual user satisfaction. 

\subsubsection{User Simulation}
\label{subsec:simulation}
User Simulators (US) are tools that are designed to simulate the user's behaviour. There are two main applications for US: i) for training the dialogue manager in an offline environment, and ii) to evaluate the dialogue policy. 

\paragraph{Training Environment.}
User Simulations are used as a learning environment to train reinforcement~-learning based dialogue managers.  They mitigate the problem of recruiting humans to interact with the systems, which is both time- and cost-intensive. There is a vast amount of literature on designing User Simulations as training environment, for a comprehensive survey refer to \cite{schatzmann:2006:us_survey}. There are several considerations to be made when building a User Simulation. 
\begin{itemize}
\item Interaction level: Does the interaction take place at the semantic level (i.e. on the level of dialogue acts) or at the surface level (i.e. using natural language understanding and generation)?
\item User goal: Does the simulation update the goal during the conversation or not? The dialogues in the second Dialogue State Tracking Challenge (DSTC2) data  contain a large amount of examples where the user changes their goal during the interaction \citep{henderson2014dstc2}. Thus, it is more realistic to model these changes as well.
\item Error model: Whether and how to realistically model the errors made by the components of the dialogue system.
\item Evaluation of the user simulation: For a discussion on this topic refer to \cite{pietquin_hastie_2013}. There are two main evaluation strategies: direct and indirect evaluation. The direct evaluation of the simulation is based on metrics (e.g. precision and recall on dialogue acts, perplexity). The indirect evaluation measures the utility of the user simulation (e.g. by evaluating the trained dialogue manager). 
\end{itemize}

%First, the level of the interaction has to be defined: does the interaction take place at the semantic level (i.e. on the level of dialogue acts) or at the word level (i.e. using natural language understanding and generation). The second consideration is concerned with the user goal; does the simulator update the goal during the conversation or not. The dialogues in the DSTC2 data \cite{henderson2014dstc2} contain a large amount of examples where the user change their goal during the interaction. Thus, it is more realistic to model these changes as well. The third consideration is on the error model, which should realistically model the errors made by the automatic speech recognition system. The last important consideration is the evaluation of the user simulation itself, for a discussion on this topic refer to \cite{pietquin_hastie_2013}. There are two main evaluation strategies: direct and indirect evaluation. The direct evaluation of the simulation are based on metrics (e.g. precision and recall on dialogue acts, perplexity). The indirect evaluation measure the utility of the user simulation (e.g. by evaluating the trained dialogue manager). 

The most popular approach to user simulation is based on the agenda-based user simulation (ABUS) \citep{Schatzmann:2007:AUS:1614108.1614146}. The simulations takes place at the semantic level, the user goal stays fixed throughout the interaction, and the user behaviour is represented as a priority ordered stack of necessary user actions. The ABUS was evaluated using indirect methods, by performing a human study on a dialogue system trained with the ABUS. The results show that the DS achieved an average task success rate of $90.6\%$ based on 160 dialogues. The ABUS system works by randomly generating a hidden user goal (i.e. the goal is unknown to the dialogue system), which consists of constraints and request slots. From this goal, the ABUS system generates a stack of dialogue acts in order to reach the goal, which is the agenda. During the interaction with the dialogue system, the ABUS adapts the stack after each turn, e.g. if the dialogue system misunderstood something, the ABUS system pushes a negation act onto the stack. 
 
Similar to other aspects of dialogue systems, more recent work is based on neural network based approaches. The Neural User Simulator (NUS) by \citep{kreyssig2018neural} proposes an end-to-end trainable architecture based on neural networks. The system performs the interaction on the surface instead of the semantic level, during the training it considers variable user goals, and the evaluation is performed indirectly. The indirect evaluation is performed from two different perspectives. First, the dialogue system, which is trained with the NUS is compared to a dialogue system trained with ABUS in the context of a human evaluation. Here, the authors report the average reward and the success rate. In both cases the NUS-trained system performs significantly better. The second evaluation is performed in a cross-model evaluation \citep{schatztnann2005effects}, i.e. the NUS-trained dialogue system is evaluated using the ABUS system and vice-versa. Here, the NUS system performed significantly better as well. This indicates that the NUS system is diverse and realistic.

\paragraph{Model Based Evaluation.}
The idea of model based evaluation is to model the user behaviour but to put more emphasis on modelling a large variety of behavioural aspects. Here, the focus does not lie in the shaping of rewards for reinforcement learning, rather, the focus lies on understanding the effects of different types of behaviour on the quality of the interaction. Furthermore, the goal is to gain insights on the effects of adapting a dialogue strategy, i.e. evaluate the changes made to the dialogue system. 
\cite{Engelbrecht:2009:ANS:1613330.1613379} introduced the MeMo workbench, which allows the modelling of user simulations. The main focus is to model different types of users and typical errors the users make. \cite{moller2006memo} introduced various types of conceptual errors, which users tend to make. There errors 
arise from the discrepancy between how the user expects the system to behave and the actual system behaviour. For instance: 
\begin{itemize}
\item State errors arise when the user input cannot be interpreted in the current state, but might be interpretable in a different state.
\item Capability errors arise when the system cannot execute the user's commands due to missing capability.
\item Modelling errors arise due to discrepancies in how the user and the system model the world. For instance, when presented with a list of options and the system allows to address the elements in the list by their positions, but the user addresses them by their name. 
\end{itemize}
On the other hand, the workbench allows the definition of various user groups based on different characteristics of a user. The characteristics used in \cite{Engelbrecht:2009:ANS:1613330.1613379} include: affinity to technology, anxiety, problem solving strategy, domain expertise, age and deficits (e.g. hearing impairment). Behavioural rules are associated to each of the characteristics. For instance, a user with high domain expertise might use a more specific vocabulary. The rules are manually curated and are engineered to influence the probabilities of user actions. 
During the interaction, the user model selects a task to solve similar to the aforementioned approaches for reinforcement-learning environments. 
In order to evaluate the user simulation, the authors compared the results of an experiment conducted with real users to the experiments conducted with the MeMo workbench. This evaluation procedure is aimed at finding whether the simulation yields the same insights as a user study. For this, they invited users from two user groups, namely older and younger users. The participants interacted with two versions of a smart-home device control system: the versions differed in the way they provide help to the users. The comparison between the user simulation and the user study results was done at various levels:
\begin{itemize}
\item High-level features, such as concept error rates or average number of semantic concepts per user turn ($\#$) AVP. Here, the results show that the simulation was not always able to recreate the absolute values, it was able to replicate the relative results. This is helpful, as it would lead to the same conclusions for the same questions. 
\item User judgment prediction which is based on a predictive model trained using the PARADISE framework. Here, the authors compared the real user judgments to the predicted judgments (where the linear model predicted the judgments of the simulated dialogue). Again, the results show that the user model would yield the same conclusions as the user study, namely that young users rated the system higher than the older users and that old users judged the dynamic help system worse than the other. 
\item Precision and Recall of predicted actions. Here, the simulation is used to predict the next user action for a given context from a dialogue corpus. The predicted user action is compared to the real user action and based on this precision and recall is computed. The results show that precision and recall are relatively low. 
\end{itemize}
The model-based user simulations are designed with the idea of allowing the evaluation of a dialogue system early in the development stage. Furthermore, they emphasize the need of interpretability, i.e. being able to understand how a certain change in the dialogue system influences the quality of the dialogue. This lies in contrast to the user simulations for reinforcement learning, which are aimed at training a dialogue system and use the reward as a measure of quality. However, the reward is often only based on the task success and the number of turns. 

\subsubsection{Subsystems Evaluation}
\label{subsec:subsyseval}
This section briefly outlines the different evaluation metrics employed on every subsystem, composing a pipelined Dialogue System, namely Natural Language Understanding, Dialogue State Tracker and Natural Language Generation systems.

\paragraph{Natural Language Understanding (NLU).} Since NLU is often cast as a classification task, NLU systems are often evaluated in the literature with regard to classification-based metrics. There are three widely used metrics~\citep{tur2011spoken}: Sentence Level Semantic Accuracy (SLSA), Slot Error Rate (SER) (also called Concept Error Rate (CER)), and F-measures. 
The SLSA measures the rate of sentences where the intents are correctly classified. The SER metric measures the rate of inserted, deleted or substituted concepts with respect to the annotated concept as a reference. 
Finally, the F-measures compute the precision and recall of the correctly detected slots.
In early systems, the distance between hypothesized sentences and reference ones is calculated with a Levenshtein distance \citep{Levenshtein_SPD66} or using the Word Error Rate~\citep{chotimongkol2001cer}, which fail to capture the semantic similarities of utterances.

\paragraph{Dialogue State Trackers (DST).} DST usually report a probability distribution over the possible next states. In order to measure the performance of such systems, accuracy and L2 metrics are widely used \citep{metallinou-bohus-williams:2013:ACL2013,henderson2014dstc2,mrk2017dst}. Accuracy measures whether the state hypothesis with the higher probability is the correct one. Having a high accuracy is crucial because DST systems must commit to a single interpretation of user's needs. L2 metric captures how well calibrated the output probabilities are, which is important when multiple dialogue states are considered in action selection.

\paragraph{Natural Language Generation (NLG).} NLG systems translate the dialogue act into natural language, the dialogue act is composed of slot-value pairs, which the NLG system renders. The evaluation focuses on two aspects: the correctness of the content and the quality of the surface realization. For the correctness, the F1 score is used \citep{mei2016talk}, as well as the slot error rate \citep{wensclstm15} (i.e. the ratio of the slots which have been correctly rendered). For the quality of the surface realization, the word overlap metrics are used (e.g. BLEU \citep{papi2002bleu}, or ROUGE \citep{lin:2004:rouge}). However, since the automated metrics do not necessarily capture all aspects of the output's quality, usually a human evaluation is performed, which usually asks about the naturalness and quality of the generated utterance \citep{dusek2020nlg}.

\section{Conversational Dialogue Systems}
\label{sec:eval_non_task}
% !TEX root = survey_dialogue_evaluation.tex
\subsection{Characteristics}
Conversational dialoge systems (also referred to as chatbots and social bots) are usually developed for unstructured, open-domain conversations with its users. They are often not developed with a specific goal in mind, other than to maintain an engaging conversation with the user~\citep{zhou2018design}. These systems are usually built with the intention to mimic human behaviour, which is traditionally assessed by the Turing Test (more on this later). However, Conversational dialogue systems might also be developed for practical applications. ``Virtual Humans", for instance, are a class of conversational agents developed for training or entertainment purposes. They mimic certain human behaviours for specific situations. For instance, a Virtual Patient mimics the behaviour of a patient, which is then used to train medical students \citep{Kenny2009virtual_human,mazza2018simulator}. Early versions of conversational agents stem from the psychology community with ELIZA \citep{Weizenbaum:1966:ECP:365153.365168} and PARRY \citep{colby_1981}. ELIZA was developed to mimic a Rogerian psychologist, whereas PARRY was developed to mimic a paranoid mind.

\paragraph{Modelling Approaches.} Generally, there are two main approaches for modelling a Conversational dialogue system: \emph{rule-based systems} and \emph{corpus-based systems}. 

Early systems, such as \emph{ELIZA} \citep{Weizenbaum:1966:ECP:365153.365168} and \emph{PARRY} \citep{colby_1981} are based on a set of rules which determine their behaviour. ELIZA works on pattern recognition and transformation rules, which take the user's input and apply transformations to it in order to generate responses. 

Recently, conversational dialogue systems have gained a renewed attention in the research community, as shown by the recent effort to generate and collect data for the (RE-)WOCHAT workshops\footnote{See \url{http://workshop.colips.org/re-wochat/} and \url{http://workshop.colips.org/wochat/}}. This renewed  attention is  motivated by the opportunity  of exploiting large amounts of dialogue data (see~\cite{serban2015surveyJournal} for an extensive study as well as Section~\ref{sec:datasets}) to automatically  author a dialogue strategy that can be used in conversational systems such as chatbots~\citep{banchs_iris:_2012,Charras2016a}. Most recent approaches train conversational agents in and end-to-end fashion using deep neural networks, which mostly rely on the sequence-to-sequence architecture~\citep{Sutskever2014seq2seq}. 
 
In the following, we focus on the corpus-based approaches used to model conversational agents. First, we describe the general concepts, and then the technologies used to implement conversational agents. Finally, we cover the various evaluation methods which have been developed in the research community. 
\subsection{Modelling Conversational Dialogue Systems}
Generally, there are two different strategies to exploit large amounts of data:
\begin{itemize}
\item Utterance Selection: Here, the dialogue is modelled as an information retrieval task. A set of candidate utterances is ranked by relevance. The dialogue structure is thus defined by the utterances in a dialogue database~\citep{lee_example-based_2009}. The idea is to retrieve the most relevant answer to a given utterance, thus learning to map multiple semantically equivalent user-utterances to an appropriate answer.  
\item Generative Models: Here, the dialogue systems are based on deep neural networks, which are trained to generate the most likely response to a given conversation history. Usually, the dialogue structure is learned from a large corpus of dialogues. Thus, the corpus defines the dialogue behaviour of the conversational agent.  
\end{itemize}
Utterance selection methods can be interpreted as an approximation to generative methods. This approach is often used for modelling the dialogue system of Virtual Humans. Usually, the dialogue database is manually curated and the dialogue system is trained to map different utterances of the same meaning to the same response utterance. Another application of utterance selection is applied to integrate different systems \citep{serban2017deep,zhou2018design}. Here, the utterance selection system selects from a candidate list, which is comprised of outputs of different subsystems. Thus, given a set of dialogue systems, the utterance selection module is trained to select for the given context, the most suitable output from the various dialogue systems. This approach is especially interesting for dialogue systems, which work on a large number of domains and incorporate a large amount of skills (e.g. set alarm clock, report the news, return the current weather forecast). 
Here, we present the technologies for corpus-based approaches, namely the neural generative models and the utterance selection models.

%Here, we present the technologies for corpus-based approaches.  The majority of the recently proposed systems are based on deep neural network approaches, where the user-utterances are encoded into a latent space. There are basically two strategies: 
%\begin{itemize}
%\item A decoder, which is conditioned on the latent representation of the user utterance, generates the response utterance. 
%\item All the candidate utterances are encoded as well and a binary classifier is used to classify the pair of user-utterance and candidate utterance as either being relevant or not. 
%\end{itemize}

\subsubsection{Neural Generative Models}
The architectures are inspired by the machine translation literature \citep{Ritter:2011:DRG:2145432.2145500}, especially neural machine translation. Neural machine translation models are based on the Sequence to Sequence (seq2seq) architecture \citep{Sutskever2014seq2seq}, which is composed of an encoder and a decoder. They are usually based on a Recurrent Neural Network (RNN). The encoder maps the input into a latent representation on which the decoder is conditioned. Usually, the latent representation of the encoder is used as the initial state of the recurrent cell in the decoder. The earliest approaches were proposed by \cite{P15-1152,vinyals2015neural}, which trained a seq2seq model on a large amount of dialogue data (in the order of $10^6$ exchanges). There are two fundamental weaknesses with the neural conversational agents. Firstly, they do not take into account the context of the conversation. Since the encoder only reads the current user input, all previous states are ignored. This leads to dialogues, where the dialogue system does not refer to previous information, which might lead to nonsensical dialogues. Secondly, the models tend to generate generic answers that follow the most common pattern in the corpus. This renders the dialogue monotonous and in the worst case leads to repeating the same answer, regardless of the current input. We briefly discuss these two aspects in the following section. 

\begin{figure}
  \begin{center}
    \includegraphics[width=0.75\textwidth]{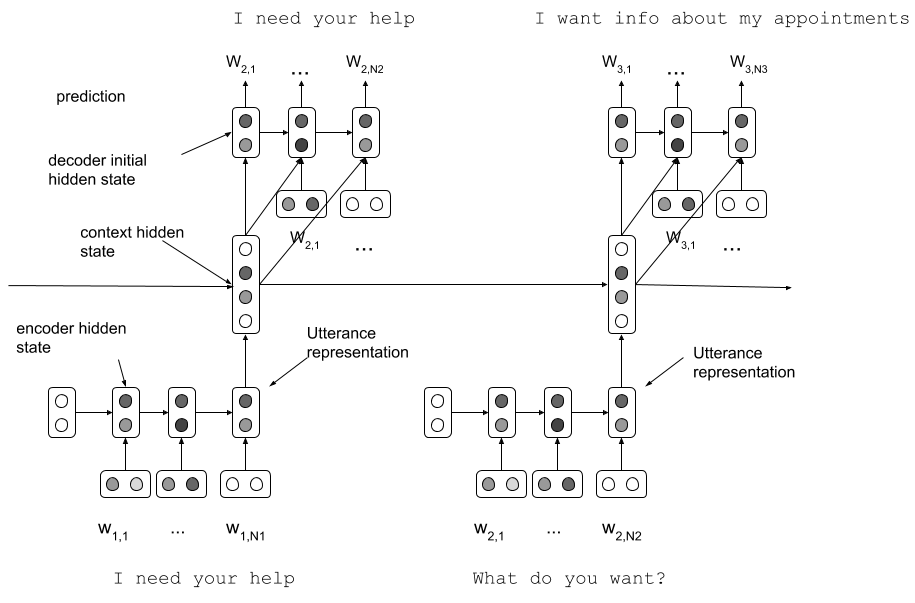}
    \caption{\label{fig:hred_overview} Overview of the HRED architecture. There are two levels of encoding: (i) the utterance encoder, which encodes a single utterance and (ii) the context encoder, which encodes the sequence of utterance encodings. The decoder is conditioned on the context encoding.}
  \end{center}
\end{figure}

\paragraph{Context.} The context of the conversation is usually defined as the previous turns in the conversations. It is important to take these into account as they contain information relevant to the current conversation.  \cite{sordoni2015contect} proposed to model the context by adding the dialogue history as a bag-of-words representation. The decoder is then conditioned on the encoded user utterance and the context representation. An alternative approach was proposed by \cite{Serban2016Hier}, who proposed the hierarchical-encoder decoder architecture (HRED), shown in Figure \ref{fig:hred_overview}, which works in three steps:
\begin{enumerate}
\item A turn-encoder (usually a recurrent neural network) encodes each of the previous utterances in the dialogue history, including the last user utterance. Thus, for each of the preceding turns a latent representation is created. 
\item A context-encoder (a recurrent neural network) takes the latent turn representations as input and generates a context representation. 
\item The decoder is conditioned on the latent context representation and generates the final output.
\end{enumerate}

The HRED architecture is used as basis for more complex neural architectures for dialogue system, such as the multi-resolution recurrent neural network (MrRNN) \citep{serban2017mrrnn}, which extends the HRED architecture by adding encoders that capture different levels of granularity (e.g. entity level, word level, or action level). Furthermore, the HRED encoder is used to generate the representation for the context in the \emph{utterance selection models} (see Section \ref{sec:utt_sel_models}).

\paragraph{Variability.} 
There are two main approaches on dealing with the issue of repetitive and universal responses: 
\begin{itemize}
\item Adapt the loss functions. The main idea is to adapt the loss function in order to penalize generic responses and promote more diverse responses. \cite{li2016diversity} propose two loss functions based on maximum mutual information: one is based on an anti-language model, which penalizes high-frequency words; the other is based on the probability of the source given the target. \cite{D16-1127} propose to train the neural conversational agent using the reinforcement-learning framework. This allows to learn a policy that can plan in advance and generate more meaningful responses. The major focus is the reward function, which encapsulates various aspects: ease of answering (reduce the likelihood of producing a dull response), information flow (penalize answers that are semantically similar to a previous answer given), and semantic coherence (based on the mutual information). 
\item Condition the decoder. The seq2seq models perform a \emph{shallow} generation process. This means that each sampled word is only conditioned on the previously sampled words. There are two methods for conditioning the generation process: condition on stochastic latent variables or on topics.  \cite{serban2017latentdialog} enhance the HRED model with stochastic latent variables at the utterance level and on the word level. At the decoding stage, first the latent variable is sampled from a multivariate normal distribution and then the output sequence is generated. \cite{xing2017topic} add a topic-attention mechanism in their generation architecture, which takes as inputs \emph{topic words} which are extracted using the Twitter LDA model \citep{Zhao:2011:CTT:1996889.1996934}. The work by \cite{ghazvininejad2017knowledge} extends the seq2seq model with a \emph{Facts Encoder}. The ``facts'' are represented as a large collection of raw texts (Wikipedia, Amazon reviews, etc.), which are indexed by named entities.
\end{itemize}

\subsubsection{Utterance Selection Methods}
\label{sec:utt_sel_models}
Utterance selection methods generally try to devise a similarity measure that measures the similarity between the dialogue history and the candidate utterances. There are roughly three different types of such measures: 
\begin{itemize}
\item Surface form similarity. This measures the similarity at the token level. This includes measures such as: Levenshtein distance, METEOR \citep{lavie2009meteor}, or TF-IDF retrieval models~\citep{Charras2016a,DubuissonDuplessis2016a}. For instance, ~\cite{PubLIMSI-6379} propose an approach that exploits recurrent surface text patterns to represent dialogue utterances.
\item Multi-class classification task. These methods model the selection task as a multi-class classification problem, where each candidate response is a single class. For instance, \cite{gandhe_surface_2013} model each utterance as a separate class, and the training data consists of utterance-context pairs on which features are extracted. Then a perceptron model is trained to select the most appropriate response utterance. This approach is suitable for applications with a small amount ($\sim 100$) of candidate answers. 
\item Neural network based approaches. Neural network architectures were introduced to leverage large amounts of training data. Usually, they are based on a siamese architecture, where both the current utterance and a candidate response are encoded. Based on this representation a binary classifier is trained to distinguish between relevant responses and irrelevant. One well-known example is the dual encoder architecture proposed by \cite{lowe2017dialog-ubuntu}. Dual Encoders transform the user input and a candidate response into a distributed representation. Based on the two representations a logistic regression layer is trained to classify the pair of utterance and candidate response as either relevant or not. The softmax score of the relevant class is used to sort the candidate responses. The authors experimented with different neural network architectures for modelling the encoder, such as recurrent neural networks or long short-term memory networks (LSTM) \citep{hochreiter1997}.
\end{itemize}

\subsection{Evaluation Methods}
\label{sec:conv_eval}
Automatically evaluating conversational dialogue systems is an open problem. The difficulty in automating this step can be attributed to the characteristics of the conversational dialogue system. Without a clearly defined goal or task to solve, and a lack of structure in the dialogues, it is not clear which attributes of the conversation are relevant to measure the system's quality. Two common approaches to assess the quality of a conversational dialogue system are to measure the appropriateness of its responses, or to measure the human likeness thereof. Both these approaches are very coarse-grained and might not reveal the complete picture. Nevertheless, most approaches in evaluation follow these principles. Depending on the characteristics of a specific dialogue system, more fine-grained approaches to evaluation can be applied, which measure the capability of the specific characteristic. For instance, a system built to increase the variability of its answers might be evaluated based on lexical complexity measures (such as token-type ratio or lexical density. For a more in-depth discussion please refer to \cite{lu2012relationship}). In the following, we introduce the automated approaches for evaluating conversational dialogue systems. In the first part, we discuss the general metrics that can be applied to both the generative models as well as the selection-based models. We then survey the approaches specifically designed for the utterance selection approaches, as they can exploit various metrics from information retrieval. 

% !TEX root = survey_dialogue_evaluation.tex
\subsubsection{General Metrics for Conversational Dialogue Systems}
\label{sec:general_metrics}
There are generally two levels in order to evaluate a conversational dialogue system: coarse-grained and fine-grained evaluations. The coarse-grained evaluations focus on the adequacy of the responses generated or selected by the dialogue system. On the other hand, fine-grained evaluations focus on specific aspects of its behaviour. 
Coarse-grained evaluations are based on two concepts: adequacy (or appropriateness) of a response, and the human likeness thereof. 
%The latter was proposed as the Turing Test \cite{TuringTest}, which is the classic procedure for evaluating a systems capability of mimicking human behaviour (and thus, be regarded as "truly" intelligent). The main concept is to test if a system can fool real humans into believing that it is actually a human.  
Fine-grained evaluations focus on specific behaviours that a dialogue system should manifest. Here, we focus on the methods devised for coherence and the ability of maintaining the topic of a conversation. 
In the following, we give an overview of the methods that have been designed to automatically evaluate the above dimensions. 

%Generally, the evaluation concepts proposed in the literature follow the concepts of adequacy of the response \cite{Lowe2017AutoTuring}, the coherence of the utterance \cite{higashinaka2014evaluating}, the ability to handle the topic of the conversation \cite{guo2018topic}, and the human-likeliness based on the Turing Test \cite{TuringTest,Bruni2017Adversarial}, and metrics based on the user experience \cite{venkatesh2018evaluating}.

%The first large scale evaluation of conversational dialogue systems was conducted in the context of the Alexa-Prize competition. Alexa Prize allowed the research community to evaluate their dialogue system live by millions of users, where the users. During this challenge the organizers collected millions of conversations, tens of millions of turns, and hundreds of thousands rated dialogues. This allowed them to conduct a large-scale study to evaluate different metrics on their correlation to human judgments \cite{venkatesh2018evaluating}. 
%In their final analysis the authors posit that the evaluation of a dialogue system has to be performed on a fine-grained level, taking into account different aspects. 

\paragraph{Appropriateness.} This is a coarse-grained concept to evaluate a dialogue, as it encapsulates many finer-grained concepts, e.g. coherence, relevance, or correctness, among others. There are two main approaches in the literature: word-overlap based metrics and methods based on predictive models inspired by the PARADISE framework (see Section \ref{sce:user_satisfaction_modelling}).
\begin{itemize}
\item \emph{Word-overlap metrics.} These metrics were originally proposed by the machine translation and the summarization community. They were initially a popular choice of metrics for evaluating dialogue systems seeing as they are easily applicable. Popular metrics such as BLEU score \citep{papi2002bleu} and ROUGE \citep{lin:2004:rouge} were used as approximation for the appropriateness of an utterance. However, \cite{liu2016eval} showed that neither of the word-overlap based scores have any correlation to human judgments.\\

Based on the criticism of the word-overlap metrics, several new metrics have been proposed. \cite{galley2015dbleu} propose to include human judgments into the BLEU score, which they call $\Delta$BLEU. The human judges rated the reference responses of the test set according to the relevance to the context. The ratings are used to weight the BLEU score to reward high-rated responses and penalize low-rated responses. The correlation to human judgments was measured by means of Spearman's $\rho$. $\Delta$BLEU has a correlation of $\rho = 0.484$, which is significantly higher than the correlation of the BLEU score, which lies at $\rho = 0.318$. Although this increases  the correlation of the metric to the human judgments, this procedure involves human judgments to label the reference sentences. 
\item \emph{Trained metrics.} \cite{Lowe2017AutoTuring} present an automatic dialogue evaluation model (ADEM), a recurrent neural network trained to predict appropriateness ratings by human judges. The human ratings were collected via Amazon Mechanical Turk, where the judges were presented with a dialogue context and a candidate response, which they rated on appropriateness on a scale from 1 to 5. Based on the ratings, a recurrent neural network was trained to score the model response, given the context and the reference response. The Pearson's correlation between ADEM and the human judgments is computed on two levels: the utterance level and at the system level, where the system level rating is computed as the average score at the utterance-level achieved by the system.\\
The Pearson's correlation for ADEM lies at $0.41$ on the utterance level and at $0.954$ on the system level. For comparison, the correlation to human judgments for the ROUGE score only lies at $0.062$ on the utterance level and at $0.268$ at the system level. \\
While ADEM relies on human labelled data, \cite{tao2018ruber} present a method, which has no need of human judges. Their model is based on two observations. Firstly, a response that is close to the ground truth is likely to be good. Secondly, a response that is related to the last utterance or the context of the conversation is good. They propose two submodels to capture these insights. The first model computes a representation of both the ground truth and the generated response based on min- and max-pooling of word embeddings. Then the cosine similarity is computed to measure the relatedness of the ground truth and the generated response. The second model rates the relatedness between the conversational context and the generated response. In order to train this model, the authors create a training set of positive examples (i.e. pairs of contexts and responses that are relevant) and negative examples (i.e. pairs of irrelevant contexts and responses). The positive examples are taken from the dialogues in the training material, whereas the negative examples are constructed by randomly sampling utterances from the corpus for a given context. Then a siamese neural network is trained on this training data to predict if a pair of context and response are relevant. The scores of both submodules are then normalized and averaged. The Pearson's correlation for their model lies at $0.4594$, which is comparable to ADEM. 

Although trained metrics have a significantly higher correlation to human judgements, they are show not to be robust \citep{sai2019re}. In fact, with simple manipulations of the response under consideration can lead to significant changes in the score of ADEM. For instance, in $48.66\%$ of cases the predicted score increased when the generated response was reversed. In $86.93\%$ of cases the predicted score increased when the generated response was replaced with a dull dummy response. Thus, creating reliable trained metrics is still an open problem.
\end{itemize}

\paragraph{Human Likeness.}
The classic approach to measure the quality of a conversational agent is the Turing Test devised by \cite{TuringTest}. The idea is to measure if the conversational dialogue system is capable of fooling a human into thinking that it is a human as well. Thus, according to this test, the main measure is the ability to imitate human behaviour. \\

Inspired by this idea, the use of \emph{adversarial learning} \citep{Goodfellow2014GAN} can be applied to evaluate a dialogue system. The framework of a generative adversarial model is composed of two parts: the generator, which generates data, and the discriminator, which tries to distinguish whether the data is real or artificially generated. The two components are trained in an adversarial manner: the generator tries to fool the discriminator, and the discriminator learns at the same time to identify if the data is real or artificial. Adversarial Evaluation of dialogue systems was first studied by \cite{kannan2017adversarial}, where the authors trained a generative adversarial network (GAN) on dialogue data, and used the performance of the discriminator as indicator for the quality of the dialogue. The discriminator achieved an accuracy of 62.5\% which indicates a weak generator. However, the authors did not evaluate whether the discriminator score is a viable metric for evaluating a dialogue system. 

A study on the viability of adversarial evaluation was conducted by \cite{Bruni2017Adversarial}. For this, they compared the performance of discriminators to the performance of humans on the task of discriminating between real and artificially generated dialogue excerpts. Three different domains were used, namely: MovieTriples (46k dialogue passages) \citep{Serban2016Hier}, SubTle (3.2M dialogue passages) \citep{Banchs2012SubTle} and Switchboard (77k dialogue passages) \citep{Godfrey1992Switch}. The GAN was trained on the concatenation of the three datasets. The evaluation was conducted on 900 dialogue passages, 300 per dataset, which were rated by humans as real or artificially generated. The results show that the annotator agreement among humans was low, with a Fleiss \citep{fleiss1971measuring} $\pi = 0.3$, which shows that the task is difficult. The agreement between the discriminator and the humans is on par with the agreement among the humans, except for the Switchboard corpus, where $\pi = 0.07$. Human annotators achieve an accuracy score with respect to the ground-truth of $64\% - 67.7\%$ depending on the domain. The discriminator achieves lower accuracy scores on the Switchboard dataset but higher scores than humans on the other two datasets. 

In order to evaluate the ability of the discriminators on different models, a seq2seq model was trained on the OpenSubtitles dataset \citep{Tiedemann285838} (80M dialogue passages). The discriminator and the human performance on the dialogues generated by the seq2seq model was evaluated. The results show that the discriminator performs better than the humans, which the authors attribute to the fact that the discriminators may pick up on patterns that are not apparent to humans. The agreement between humans and the discriminator is very low. 

\paragraph{Fine-grained Metrics.}
The above methods for evaluating conversational dialogue systems work on a coarse-grained level. The dialogue is evaluated on the basis of producing adequate responses or its ability to emulate human behaviour. These concepts encompass more finer-grained concepts. In this section, we look at topic-based evaluation. 

\emph{Topic-based evaluation}. This measures the ability of a conversational agent to talk about different topics in a cohesive manner. \cite{guo2018topic} propose two dimensions of topic-based evaluation: topic breadth (can the system talk about a large variety of topics?) and topic depth (can the system sustain a long and cohesive conversation about one topic?). For topic classification, a Deep Averaging Network (DAN) was trained on a large amount of question data. DANs do topic classification and the detection of topic-specific keywords. The conversational data used to evaluate the topic-based metrics stems from the Alexa-Prize challenge~\footnote{\url{https://developer.amazon.com/alexaprize}}, which consists of millions of dialogues and hundreds of thousands of live user ratings (on a scale from 1 to 5). Using the DAN, the authors classified the dialogue utterances according to the topics.

Conversational \emph{topic depth} is measured by the average length of a sub-conversation on a specific topic, i.e. multiple consecutive turns where the utterances are classified as being the same topic. The conversational breadth is measured on a coarse- and fine-grained level. Coarse-grained topic breadth is measured as the average number of topics a bot converses about during a conversation. On the other hand, \emph{topic breadth} measures looks at the total number of distinct topic keywords across all conversations.

To measure the validity of the proposed metrics, correlations between the metric and the human judgments are computed. The conversational topic depth metric has a correlation of $\rho = 0.707$ with the human judgments. The topic breadth metric has a correlation of  $\rho = 0.512$ with the human judgments. The lower correlation of the topic breadth is attributed to the fact that the users may not have noticed a bot repeating itself as they only conversed with a bot a few times.

%!TEX root = survey_dialogue_evaluation.tex
\subsubsection{Utterance Selection Metrics}
\label{subsec:utterance-selection}
The evaluation of dialogue systems based on utterance selection differs from the evaluation of generation-based dialogue systems. Here, the evaluation is based on metrics used in information retrieval, especially Recall@k (R@k). R@k measures the percentage of relevant utterances among the top-k selected utterances. One major drawback of this approach is that potentially correct utterances among the candidates could be regarded as incorrect. 

\paragraph{Next Utterance Selection.} \cite{lowe2016nuc} evaluate the impact of this limitation and evaluate whether the Next Utterance Classification (NUC) task is suitable to evaluate dialogue systems. For this, they invited 145 participants from Amazon Mechanical Turk (AMT) and 8 experts from their lab. The task was to select the correct response given a dialogue context (of at most six turns) and five candidate utterances, of which exactly one is correct. Note that the other four utterances could also be relevant, but are regarded as incorrect in this experiment. The study was performed on dialogues of three different domains: the SubTle Corpus \citep{Banchs2012SubTle} consisting of movie dialogues, the Twitter Corpus \citep{Ritter:2010:TwitterCorpus} consisting of user dialogues, and the Ubuntu Dialogue Corpus \citep{Lowe2015TheUD}, which consists of conversations about Ubuntu related topics. 

The human performance was compared to the performance of an artificial neural network, which is trained to solve the same task. The performance was measured by means of R@1 score. The results show that for all domains, the human performance was significantly above random, which indicates that the task is feasible. Furthermore, the results show that the human performance varies depending on the domain and the expertise level. In fact, the lab participants performed significantly better on the Ubuntu domain, which is regarded as harder as it requires expert knowledge. This shows that there is a range of performance that can be achieved. Finally, the results showed that the ANN achieved similar performance to the human non-experts and performed worse than the experts. This shows that this task is not trivial and by far not solved. However, the authors did not take into account the fact that multiple candidates responses could be regarded as correct. This is possible since the selection of the candidate response is performed by sampling at random from the corpus. On the other hand, it is not clear if their evaluation suffered from this potential limitation, as their results showed the feasibility and relevance of the NUC task.

\cite{DeVault:2011:TLE:2132890.2132896} and \cite{Gandhe2016} tackle the problem of having multiple relevant candidate utterances and propose a metric which takes this into account. Their metrics are both dependent on human judges and measure the appropriateness of an utterance. 

\paragraph{Weak Agreement.} \cite{DeVault:2011:TLE:2132890.2132896} propose the \emph{weak agreement} metric. This metric is based on the observation that human judges only agree in about 50\% of the cases on the same utterance for a given context. The authors attribute this to the fact that multiple utterances could be regarded as acceptable choices. Thus, the weak agreement metric regards an utterance as appropriate if at least one annotator chose this utterance to be appropriate. 

The authors apply the weak agreement metric on the evaluation of a Virtual Human which simulates a witness in a war-zone and is designed to train military personnel in Tactical Questioning \citep{gandhe2009integrated}. They gathered 19 dialogues and 296 utterances in a Wizard-of-Oz experiment. To allow for more diversity, they let human experts write paraphrases of the commander role to ensure that the virtual character understands a larger variety of inputs. Furthermore, the experts expanded the set of possible answers by the virtual character by annotating other candidate utterances as appropriate. 

The weak agreement metric was able to measure the improvement of the system when the extended dataset was applied: the simple system based on the raw Wizard-of-Oz data achieved a weak agreement of 43\%; augmented with the paraphrases, the system achieved a score of 56\%; and, finally, adding the manual annotation increases the score to 67\%. Thus, the metric is able to measure the improvements made by the variety in the data.

\paragraph{Voted Appropriateness.} One major drawback of the weak agreement is that it depends on human annotations and is not applicable to large amounts of data. \cite{Gandhe2016} improve upon the idea of weak agreement by introducing the \emph{Voted Appropriateness} metric. Voted Appropriateness takes the number of judges into account which selected an utterance for a given context. In contrast to weak agreement, which regarded each adequate utterance equally, Voted Appropriateness weights each utterance. 

Similarly to the PARADISE approach, the authors of Voted Appropriateness fit a linear regression model on the pairs of utterances and contexts labelled with the amount of judges that selected the utterance. The fitted model only explains 23.8\% of the variance. The authors compared the correlation of the Voted Appropriateness and the weak agreement metric to human judgments. The correlation was computed on the individual utterance level and the system level. For the system level, the authors used data from seven different dialogue systems and averaged the ratings over all dialogues of one system. On the interaction level, the Voted Appropriateness achieved a correlation score of 0.479 $(p < 0.001,\ n=397)$, and the weak agreement achieved 0.485 $(p < 0.001,\ n=397)$. On the system level, Voted Appropriateness achieved 0.893 $(p < 0.01,\ n=7)$ and weak agreement achieved 0.803 $(p < 0.001,\ n=397)$. Thus, on the system level Voted Appropriateness performs closer to human judgments. Both metrics rely heavily on human annotations, which makes the metrics hardly suitable for large-scale data driven approaches.

\section{Question Answering Dialogue Systems}
\label{sec:eva_qa_dialogue}
% !TEX root = survey_dialogue_evaluation.tex

A different form of task-oriented systems are Question Answering (QA) systems. Here, the task is defined as finding the correct answer to a question. This setting differs from the aforementioned task-oriented systems in the following ways:
\begin{itemize}
\item Task-oriented systems are developed for a multitude of tasks (e.g. restaurant reservation, travel information system, virtual assistant, etc.),whereas QA systems are developed to find answers to specific questions.
\item Task-oriented systems are usually domain-specific, i.e. the domain is defined in advance through an ontology and remains fixed. In contrast, QA systems usually work on broader domains (e.g. factoid QA can be done over different domains at once), although there are also some QA systems focused only on a specific domain \citep{Sarrouti2017,DoNTNN17}.
\item The dialogue aspect for QA systems is not tailored to sound human-like, rather, the focus is set on the completion of the task. That is, to provide a correct answer to the input question.
\end{itemize}

\subsection{Characteristics}
Generally, QA systems allow the users to search for information using a natural language interface, and return short answers to the user's question \citep{Voorhees2006}. QA systems can be broadly categorized into three categories \citep{bernardi10from}: single-turn QA, context QA, and Interactive QA.

\paragraph{Single-turn QA.} Single-turn QA  is the most common type of system. Here, the system is developed to return a single answer to the users' question without any further interaction. These systems work very well for factoid questions \citep{Voorhees2006}. However, they have difficulties handling complex questions, which require several inference steps \citep{P17-1167} or situations where systems need additional information from the user \citep{Li:2017:CAN:3097983.3098115}.

Single-turn QA can be approached from two main perspectives \citep{Rogers2020quail}:

\begin{itemize}
    \item Open QA, where systems collect evidences and answers across several sources such as Web pages and knowledge bases \citep{fader-etal-2013-paraphrase}
    \item Reading Comprehension (RC), where the answer is gathered from a single document. This is the most common approach.
\end{itemize}

RC systems can be oriented to:

\begin{itemize}
    \item  Extractive RC, where systems extract spans of text with the answer. This approach has received a lot of attention fostered by the availability of popular benchmarks such as SQuAD \citep{Rajpurkar2018}, NewsQA \citep{trischler-etal-2017-newsqa} or TriviaQA \citep{joshi-etal-2017-triviaqa}. Each of these datasets contains thousands of examples, which permits to train Deep Learning systems and obtain good results. 
    
    \item Multiple-choice RC, where systems must select an answer from a set of candidates. Multiple-Choice (MC) is a common way to measure reading comprehension in humans. This is why some researches have pointed MC as a better format to test language understanding of automatic systems \citep{Rogers2020quail}. There exists several MC collections, mostly in English. In some cases it involves paying crowd-workers to gather documents and/or pose questions regarding those documents. MCTest \citep{Richardson2013}, for example, proposed for the workers to invent short, children friendly, fictional stories and four questions with four answers each, including deliberately wrong answers. As a way to encourage a deeper understanding of texts, the QuAIL dataset includes unanswerable questions \citep{Rogers2020}. Other datasets were created from real world exams. This is the case of the well known MC dataset RACE \citep{Lai2017}, or the multilingual Entrance Exams \citep{rodrigo_systems_2018}.
    
    \item Generative QA, where systems create a text that answers the question. The exact text is not necessarily contained in any document, which makes this a challenging task. This kind of systems has received less attention given that it is difficult to perform an exact evaluation and there are few datasets available \citep{kocisky-etal-2018-narrativeqa}.
\end{itemize}

There is a large amount of research in the area of single-turn QA and there are several surveys, we refer the reader to: \cite{Kolomiyets2011,Diefenbach2018,Mishra2016}. In this survey, we focus on the evaluation of multi-turn QA systems, which is a much less researched area.

\paragraph{Context QA.} Context QA  refers to systems which allow for follow-up questions to resolve ambiguities or keeping track of a sequence of inference steps \citep{PenasMFSRG12}. The questions can be highly context-dependent and elliptical, with references to previous questions and answers, which can be seen as a dialog. In fact, it is common to include pronouns instead entities. That is, systems must rely not only on the source document and last question, but also on the context given by previous questions and answers. 

Context QA  systems are also named multi-turn QA \citep{choi2018quac} or sequential QA \citep{DBLP:conf/aaai/SahaPKSC18}. The most common approach is to develop these systems for extractive RC. In some cases, context QA systems are used for answering complex questions. These systems assume that some complex questions are usually unrealistic but they can be decomposed into simpler inter-related questions \citep{iyyer-etal-2017-search}. Then, the system answers the simpler questions and obtain an answer to the initial complex question \citep{talmor-berant-2018-web}.

\paragraph{Interactive QA.} Interactive QA (IQA) systems combine context QA systems and task-oriented dialogue systems. The main purpose of the conversation module is to handle under-or-over constrained questions \citep{qu2002constraint}. E.g. if a question does not yield any results, the system might propose to relax some constraints. In contrast, if a question yields too many results, the interaction can be used to introduce new constraints to filter a list of results \citep{rieser_lemon_2009}. For a more in-depth discussion on IQA systems, refer to \cite{Konstantinova2013}.

\subsection{Technologies}

Current QA technologies for single-turn QA are based on pre-trained transformer models such as BERT   \citep{devlin-etal-2019-bert}, XLNet \citep{yang2019xlnet} or ALBERT \citep{lan2019albert}.  These models have been pre-trained from unlabeled text to do Masked Language Modeling and Next Sentence Prediction. Afterwards, each model can be fine-tuned in specific tasks such as those at Glue \citep{wang-etal-2018-glue} or QA.

Fine-tuning for QA systems is done by modelling the span detection as prediction of the start and end token in the paragraph. The input to the system is a pair of question and paragraph. Thus, the trained system will output the span with the highest probability of being an answer to the question. These systems achieve the best results for extractive QA, as it can be seen in the corresponding leaderboards of the most popular collections\footnote{\url{https://rajpurkar.github.io/SQuAD-explorer/}}

In the case of multi-turn QA, systems must be aware of the dialogue history. One approach is to reuse single-turn systems, augmenting the input with previous questions and answers \citep{huang2018flowqa}. In some cases, the system may focus on modelling information gain and include pre-trained models such as BERT \citep{yeh-chen-2019-flowdelta}.

Given the importance of dealing with answer history, other researchers have proposed to represent answer history using embeddings from pre-trained models \citep{10.1145/3357384.3357905}. Then, the system includes also a history attention mechanism to help in the selection of items in the history of the dialogue.

Other models include Adversarial Training and Knowledge Distillation over ROBERTA \citep{liu2019roberta} to perform a better fine-tunning of pre-trained models \citep{ju2019technical}. While Adversarial Training allows improving the performance of the system against data perturbations, Knowledge Distillation transfers knowledge from one machine to another to improve results of the second machine \citep{DBLP:conf/icml/FurlanelloLTIA18}.

\subsection{Evaluation of QA Dialogue Systems}
The evaluation of QA systems has two aspects: the correctness of the answer and the flow of the conversation. Currently, most QA systems are evaluated based on the correctness of their answers. Even for multi-turn QA systems, the dialogue flow is often ignored during evaluation \citep{reddy2018coqa,choi2018quac}.

\paragraph{Correctness Metrics.} The evaluation of QA systems depends on the output of the system. For open QA, where the output is a ranking of sentences with potential answers, the evaluation is mostly based on ranking measures such as Mean Average Precision (MAP) or Mean Reciprocal Rank (MRR), but there are also evaluations based on precision, recall and F1 \citep{yang-etal-2015-wikiqa}.

For multiple-choice RC the task is evaluated using accuracy \citep{Clark_Etzioni_2016}, that is, the number of times in which the system selected the correct answer.

For extractive QA, which is the most common approach, the output is a span of text. The retrieved span is compared with the ground truth answers and two kinds of evaluations are given \citep{rajpurkar-etal-2016-squad}:

\begin{itemize}
    \item Exact matching, which measures the percentage of candidate answers that match any one of the ground truth answers exactly.
    \item Approximate matching based on F1, which measures the macro-average overlap between the bag of words of candidates and ground truth answers.
\end{itemize}

\paragraph{Dialogue Evaluation.} The nature of multi-turn QA systems makes it quite hard to design accurate evaluation frameworks that go beyond the correctness measures, which do not take into consideration the dialogue aspect of the interaction.  In fact, a proper evaluation of multi-turn QA systems requires humans to interact with the systems. 
%However, the metrics used to assess the quality are often based on metrics used in  Information Retrieval (IR). For instance, \cite{Li:2017:CAN:3097983.3098115} report the error rate of the system. Also \cite{Kelly:2007:OTC:1273221.1273231} base the valuation on F-scores, which is computed over “information nuggets”. These nuggets are retrieved by the assessors of the system, and thus, this evaluation method heavily depends on human involvement. 
The first evaluation framework designed specifically for IQA systems is based on a series of questionnaires to capture different aspects of the system \citep{kelly2009questionnaires}. The authors argue that metrics based on the relevance of the answers are not sufficient to evaluate an IQA system (e.g. it does not take the user feedback into account). Thus, they evaluate the usage of different questionnaires in order to assess the different systems. The questionnaires they propose are:
\begin{itemize}
\item NASA TLX (cognitive workload questionnaire): Used to measure the cognitive workloads as subjects completed different scenarios.
\item Task Questionnaire: After each task the questionnaire is filled out, which focuses on the experiences of using a system for a specific task. 
\item 
System Questionnaire: Compiled after using a system for multiple tasks. This measures the overall experiences of the subjects. 
\end{itemize}
Their evaluation showed that the Task Questionnaire is the most effective at distinguishing among different systems.

The evaluation of dialogue QA systems requires one to simulate some interactions with users and evaluate them. These interactions can, on the one hand, be created by real users, which is associated with high costs, and makes it hard to reproduce the experiments and reuse the data. For example, \cite{li-etal-2017-dailydialog} developed DailyDialog, a multi-turn dataset with 13k dialogues created by humans that also include emotion information.

On the other hand, the interactions can be automatically produced. However, it is challenging to create users' responses automatically. One approach for creating simulations is to provide some feedback based on the supplementary questions. For example, if an additional question asks for a location, the simulator can return a location contained in the dialogue's history (or related to it) \citep{Li:2017:CAN:3097983.3098115}. Nevertheless, the simulation can generate several errors. On the other hand, the simulation might only reward the generation of questions similar to a given template \citep{D16-1127}, which constrains the diversity of questions.

There is usually a weak correlation between automatic evaluations and human judgements in multi-turn QA. This is because most of the current QA dialogue systems are trained and tested using data where there is only a single response for each context \citep{serban2017latentdialog}. Moreover, this data contains only a possible path to reach the correct answer, while the same answer could be reached with a different dialogue. In fact, there are many features involved in deciding the next response in a dialogue. This has been defined as the one-to-many problem of dialogues \citep{zhao-etal-2017-learning}. 

Automatic evaluations based on multiple-reference responses have been proposed to alleviate the one-to-many problem. Multi-reference based evaluations include several correct responses for a given context. Thus, these evaluations promote diversity better than single-response approaches.

\cite{sordoni2015contect} created a synthetic multiple-reference dialogue corpus based on Twitter. Additional responses to the initial response were searched using Information Retrieval and rated by crowd workers. The authors kept only responses with a high rate. \cite{galley2015dbleu} created a dataset from Twitter following the work from \cite{sordoni2015contect}. However, \cite{galley2015dbleu} included all the synthetic responses (no matter the rate given by crowd workers) and used the data for testing a  new metric called Discriminative BLEU. 

\cite{Sugiyama2019} performed another evaluation based on multiple-reference responses. They measured the correlation, using a regression-based approach, between systems' responses and a large set of both positive and negative human references. \cite{gupta2019investigating} extended the test split of DailyDialog (1k dialogues) with multiple references. They compared the results of using single-reference versus multiple-reference data. Both works showed a higher correlation of automatic evaluations with human judgments when using multiple-reference dialogues instead of single-reference data.

%!TEX root = survey_dialogue_evaluation.tex
\section{Evaluation Datasets and Challenges}
\label{sec:datasets}
Datasets play an important role for the evaluation of dialogue systems, together with challenges open to public participation. A large number of datasets have been used and made publicly available for the evaluation of dialogue systems in the last decades, but the coverage across dialogue components and evaluation methods (e.g. Sections \ref{sec:eval_task} and \ref{sec:eval_non_task}) is uneven. Also note that datasets are not restricted to specific evaluation methods, as they can be used to feed more than one evaluation method or metric interchangeably. In this section, we cover the most relevant datasets and challenges, starting with select datasets. For further references, refer to a broad survey of publicly available datasets that have already been used to build and evaluate dialogue systems carried out by \cite{serban2015surveyJournal}.

The dialogue datasets selected for this section are listed in Tables \ref{tab:task}, \ref{tab:conv} and \ref{tab:qa}, where properties such as the topics covered and number of dialogues are indicated.

\subsection{Datasets for Task-Oriented Dialogue Systems}

\begin{table}[t]
\centering
\begin{tabular}{llll}
\hline
Name & Topics & \# dialogues   &  Reference\\ \hline
DSTC1 & Bus schedules & 15,000  & \citep{williams2013} \\
DSTC2 & Restaurants & 3,000 & \citep{henderson2014dstc2}\\
DSTC3 & Tourist information & 2,265 & \citep{dstc2-3} \\
DSTC4 \& DSTC5 & Tourist information & 35 & \citep{kim2016dstc5}\\
DSTC6 & Restaurant reservation &-  & \citep{dstc6}\\
DSTC7 (Flex Data) & Student guiding & 500 & \citep{dstc7task1} \\
DSTC8 (MetaLWOz)& 47 domains & 37,884  & \citep{dstc8}\\
DSTC8 (Schema-Guided)& 20 domains&22,825&\citep{rastogi2019scalable}\\
MultiWOZ & Tourist information & 10,438  & \citep{budzianowski2018large}\\
Taskmaster-1 & 6 domains & 13,215  & \citep{byrne19} \\
MultiDoGo & 6 domains & 86,698  & \citep{peskov-etal-2019-multi} \\
\hline
\end{tabular}
\caption{Datasets for task-oriented dialogue systems.}
\label{tab:task}
\end{table}

Datasets are usually designed to evaluate specific dialogue components, and very few public datasets are able to evaluate an entire task-oriented dialogue system (e.g. Section \ref{sec:eval_task}). The evaluation of these kinds of systems is highly system-specific, and it is therefore difficult to reuse the dataset with other systems. Their evaluation also requires considerable human effort, as  the involvement of individual users or external evaluators is usually needed. For example, in \cite{gasic2013domain}, which is a Partially observable Markov decision process -based dialogue system mentioned in Section \ref{subsec:slot} for the restaurants domain, the evaluation of policies is done by crowd-sourcers via the Amazon Mechanical Turk service. Mechanical Turk users were asked first to find some specific restaurants, and after each dialogue was finished, they had to fill in a feedback form to indicate if the dialogue had been successful or not. Similarly, for the end-to-end dialogue system by \cite{wen2017e2e_dialog} (cf. Section \ref{subsec:ent-to-end}), also for the restaurants domain, human evaluation was conducted by users recruited via Amazon Mechanical Turk. Each evaluator had to follow a given task and to rate the system's performance. More specifically, they had to grade the subjective success rate, the perceived comprehension ability and naturalness of the responses. 

Most of the task-oriented datasets are designed to evaluate components of dialogue systems. For example, several datasets have been released through different editions of the Dialog State Tracking Challenge\footnote{\url{https://www.microsoft.com/en-us/research/event/dialog-state-tracking-challenge/}}, focused on the development and evaluation of the dialogue state tracker component. However, even if these datasets were designed to test state tracking, \cite{DBLP:journals/corr/BordesW16} used them to build and evaluate a whole dialogue system, re-adjusting the dataset by ignoring the state annotation and reusing only the transcripts of dialogues. The Schema Guided Dialogue (SGD) dataset released for the 8th edition of DSTC was designed to test not only state tracking, but also intent prediction, slot filling and language generation for large-scale virtual assistants. SGD consists of almost 23K annotated multi-domain (bank, media, calendar, travel, weather, ...), task-oriented dialogues between a human and a virtual assistant.

The MultiWOZ (Multi-Domain Wizard-of-Oz) dataset represented a significant breakthrough in the scarcity of dialogues as it contains around 10K dialogues, which is at least one order of magnitude larger than any structured corpus available before  \citep{budzianowski2018large}. It is annotated with dialogue belief states and dialogue actions, so it can be used for the development of individual components of a dialogue system. But its considerable size makes it very appropriate for the training of end-to-end based dialogue systems. The main topic of the dialogues is tourism, containing seven domains, such as attractions, hospitals, police, hotels, restaurants, taxis and trains. Each dialogue can contain more than one of these domains.

Similar in size and content to MultiWOZ is Taskmaster-1 task-based dialogue dataset \citep{byrne19}. It includes around 13K dialogues in six domains: ordering pizza, setting auto repair appointments, arranging taxi services, ordering movie tickets, ordering coffee drinks and making restaurant reservations. What makes it different from the previous one is that more than a half of the dialogues are created following a self-dialogue methodology, in which a crowd-worker writes the full dialogue themselves. The authors claim that these self-dialogues have richer and more diverse language than, for example, MultiWOZ, as it is not restricted to a small knowledge base.

The largest human-generated and multi-domain dialogue dataset that is available to the public is MultiDoGo \citep{peskov-etal-2019-multi}, which comprises over 81K dialogues. These dialogues were created following the Wizard-of-Oz approach between a crowd-worker and a trained annotator. These participants were guided to introduce specific biases like intent or slot change, multi-intent, multiple slot values, slot overfilling and slot deletion in conversations. Additionally, over 54K of the total amount of the dialogues are annotated at the turn level for intent classes and slot labels. Dialogues are from six different domains: airline, fast food, finance, insurance, media and software support.

We will conclude this section by discussing two related tools, rather than a dialogue dataset. The first tool, called PyDial\footnote{\url{http://www.camdial.org/pydial/}}, partially addresses the shortage of evaluation datasets for task-oriented systems. This is because it offers the opportunity for developing a dialogue management environment, based on reinforcement-learning for benchmarking purposes \citep{ultes2017pydial}. Thus, it makes it possible to evaluate and compare different task-oriented dialogue systems in the same conditions. This toolkit not only provides domain-independent implementations of different modules in a dialogue system, but also simulates users (see Section \ref{subsec:simulation}). It uses two metrics for the evaluation: (1) the average success rate and (2) the average reward for each evaluated policy model of reinforcement-learning algorithms. The success rate is defined as the percentage of dialogues that are completed successfully. Thus, it is closely related to the task-completion metric used by the PARADISE framework (see Section \ref{subsec:paradise}).

Another dialogue annotation tool is called LIDA \citep{collins19}. The authors argue that the quality of a dataset has a significant effect on the quality of a dialogue system, hence, a good dialogue annotation tool is essential to create the best annotated dialogue dataset. LIDA is the first annotation tool that handles the entire dialogue annotation pipeline from turn and dialogue segmentation through to labelling structured conversation data. Moreover, it also includes an interface for inter-annotator disagreements resolution.

\subsection{Datasets for Conversational Dialogue Systems}
\label{subsec:conv}

\begin{table}[t]
\centering
\begin{tabular}{llll}
\hline
Name & Topics & \# dialogues      & Reference\\ \hline
Switchboard & Casual topics & 2,400 & \citep{Godfrey1992Switch} \\
British National Corpus & Casual topics & 854 & \citep{leech1992} \\
SubTle Corpus & Movie subtitles & 3,35M & \citep{Ameixa2013FromST} \\
Reddit Domestic Abuse Corpus & Abuse help & 21,133 & \citep{schrading2015}\\
Twitter Corpus & Unrestricted & 1,3M & \citep{Ritter:2010:TwitterCorpus} \\
Twitter Triple Corpus & Unrestricted & 4,322 & \citep{sordoni2015contect} \\
Ubuntu Dialogue Corpus & Ubuntu problems  & 930K & \citep{Lowe2015TheUD} \\
bAbI & Restaurant reservation & 3,000 & \citep{DBLP:journals/corr/BordesW16} \\
OpenSubtitles & Movie subtitles & 36M & \citep{TIEDEMANN12.463} \\
CornellMovie & Movie dialogues & 220K & \citep{danescu-niculescu-mizil-lee-2011-chameleons} \\
\hline
\end{tabular}
\caption{Datasets for conversational dialogue systems.}
\label{tab:conv}
\end{table}

Regarding the evaluation of Conversational dialogue systems presented in Section \ref{sec:eval_non_task}, datasets derived from conversations on micro-blogging or social media websites (e.g. Twitter or Reddit) are good candidates, as they contain general-purpose or non-task-oriented conversations that are orders of magnitude larger than other  dialogue datasets used before. For instance, Switchboard \citep{Godfrey1992Switch} (telephone conversations on pre-specified topics), British National Corpus \citep{leech1992} (British dialogues many contexts, from formal business or government meetings to radio shows and phone-ins) and SubTle Corpus \citep{Ameixa2013FromST} (aligned interaction-response pairs from movie subtitles) are three datasets released earlier that have 2,400, 854 and 3.35M dialogues and 3M, 10M and 20M words, respectively. These sizes are relatively small if we compare to the huge Reddit Corpus\footnote{\url{https://www.reddit.com/r/datasets/comments/3bxlg7/i_have_every_publicly_available_reddit_comment/}} which contains over 1.7 billion comments\footnote{As far as we know, this dataset has not been used in any research work. Researchers have used smaller and more curated versions of the Reddit dataset like Reddit Domestic Abuse Corpus \cite{schrading2015}, which contains 21,133 dialogues.}, or the Twitter Corpus described below.

Because of the limit on the number of characters permitted in each message on Twitter, the utterances are quite short, very colloquial and chat-like. Moreover, as the conversations happen almost in real-time, the conversations of this micro-blogging website are very similar to spoken dialogues between humans. There are two publicly available large corpora extracted from Twitter. The former one is the Twitter Corpus presented in \cite{Ritter:2010:TwitterCorpus}, which contains roughly 1.3 million conversations and 125M words drawn from Twitter. The latter is a collection of 4,232 three-step (context-message-response) conversational snippets extracted from Twitter logs\footnote{\url{https://www.microsoft.com/en-us/download/details.aspx?id=52375}}. This is labeled by crowdsourced annotators, who measure the quality of a response in a given context \citep{sordoni2015contect}.

Alternatively, \cite{Lowe2015TheUD} hypothesized that chat-room style messaging is more closely correlated to human-to-human dialogues than micro-blogging websites like Twitter, or forum-based sites such as Reddit. Thus, they presented the above-mentioned Ubuntu Dialogue Corpus. This large-scale corpus targets a specific domain. Thus, it could accordingly be used as a task-oriented dataset for the research and evaluation of dialogue state trackers. However, it also has the unstructured nature of interactions from microblog services that makes it appropriate for the evaluation of non-task-oriented dialogue systems. %For example, dialogue managers based on neural language models that make use of large amounts of unlabeled data.

These two large datasets are adequate for the three subtypes of non-task-oriented dialogue systems: unsupervised, trained and utterance selection metrics. Notice that, additionally, some human judgments could be needed in some cases, such as in \cite{Lowe2017AutoTuring} for the ADEM system (see Section \ref{sec:general_metrics}). Here, they use human judgments collected via Amazon Mechanical Turk in addition to the evaluation using the Twitter dataset. 

Apart from the afore-mentioned two datasets, the five datasets generated recently for bAbI tasks \citep{DBLP:journals/corr/BordesW16} are appropriate for evaluation using the next utterance classification method (see Section \ref{subsec:utterance-selection}). These tasks were designed for testing end-to-end dialogue systems in the restaurant domain, but they check whether the systems can predict the appropriate utterances among a fixed set of candidates, and are not useful for systems that generate the utterance directly. The ibAbI dataset mentioned in the next section has been created based on bAbI to cover several representative multi-turn QA tasks.

Another interesting resource is the ParlAI framework\footnote{\url{http://parl.ai/}} for dialogue research, as it contains many popular datasets available all in one place with the goal of sharing, training and evaluating dialogue models across many tasks \citep{miller2017parlai}. Some of the dialogue datasets that are included have been already mentioned (bAbI Dialog tasks and the Ubuntu Dialog Corpus) but it also contains conversations mined from OpenSubtitles\footnote{\url{http://opus.lingfil.uu.se/OpenSubtitles.php}} and Cornell Movie\footnote{\url{https://www.cs.cornell.edu/~cristian/Cornell_Movie-Dialogs_Corpus.html}}.

\subsection{Datasets for Question Answering Dialogue Systems}

\begin{table}[t]
\centering
\begin{tabular}{llll}
\hline
Name & Topics & \# dialogues   &  Reference\\ \hline
Ubuntu Dialogue Corpus & Ubuntu problems & 930K & \citep{Lowe2015TheUD} \\
MSDialog & Microsoft products & 35K & \citep{DBLP:conf/sigir/QuYCTZQ18} \\
ibAbI & Restaurant reservation & - & \citep{Li:2017:CAN:3097983.3098115} \\
CoQA &7 domains & 8,399 & \citep{reddy2018coqa} \\
QuAC & People & 13,594 & \citep{choi2018quac} \\
DoQA & Cooking & 1,637 & \citep{campos2019} \\
\hline
\end{tabular}
\caption{Datasets for question answering dialogue systems.}
\label{tab:qa}
\end{table}

With respect to QA dialogue systems, two datasets have been created based on human interactions from technical chats or forums. The first one is the Ubuntu Dialogue Corpus, containing almost one million multi-turn dialogues extracted from the Ubuntu chat logs, which was used to receive technical support for various Ubuntu-related problems \citep{Lowe2015TheUD}. Similarly, MSDialog contains dialogues from a forum dedicated to Microsoft products. MSDialog also contains the user intent of each interaction \citep{DBLP:conf/sigir/QuYCTZQ18}.

ibAbI represents another approach for creating multi-turn QA datasets \citep{Li:2017:CAN:3097983.3098115}. ibAbI interactivity adds to the bAbI dataset that was previously presented (see Section \ref{subsec:conv}) by adding sentences and ambiguous questions with the corresponding disambiguation question, which should be asked by an automatic system. The authors evaluate their system regarding the successful tasks. However, it is unclear how to evaluate a system if it produces a modified version of the disambiguation question.

Recently, several datasets that are very relevant for the context of QA dialogue systems have been released. The CoQA (Conversational Question Answering) dataset contains 8K dialogues and 127K conversation turns \citep{reddy2018coqa}. The answers from CoQA are free-form text with their corresponding evidence highlighted in the passage. It is a multi-domain dataset, as the passages are selected from several sources, covering seven different domains: children's stories, literature, middle and high school English exams, news, articles from Wikipedia, science and discussions from Reddit. QuAC (Question Answering in Context) consists of 14K information-seeking QA dialogues (100K total QA pairs) over sections from Wikipedia articles about people \citep{choi2018quac}. What makes it different from other datasets so far is that some of the questions are unanswerable and that context is needed in order to answer some of the questions. Another similar dataset that has unanswerable questions and its questions are context-dependent is DoQA, a dataset for accessing domain-specific Frequently Asked Question sites via conversational QA \citep{campos2019}. It contains 1,637 information-seeking dialogues on the cooking domain (7,329 questions in total). An analysis carried out by the authors showed that in this dataset there are less factoid questions than in the others, as DoQA focuses on open-ended questions about specific topics. Amazon Mechanical Turk was used to collect the dialogues for the three datasets.

\subsection{Evaluation Challenges}

We conclude this section by summarizing some of the recent evaluation challenges that are popular for benchmarking state-of-the-art dialogue systems. They have an important role in the evaluation of dialogue systems, not only because they offer a good benchmark scenario to test and compare the systems on a common platform, but also because they often release the dialogue datasets for later evaluation.

Perhaps one of the most popular challenges is the Dialog State Tracking Challenge (DSTC)\footnote{\url{https://www.microsoft.com/en-us/research/event/dialog-state-tracking-challenge/}}, which was previously mentioned in this section. DSTC was started in 2013 in order to provide a common testbed for the task of dialogue state tracking. It continued on a yearly basis with remarkable success. For its sixth edition, it was renamed as Dialog System Technology Challenges due to the interest of the research community in a wider variety of dialogue-related problems. Various well-known datasets have been produced and released for every edition: DSTC1 has human-computer dialogues in the bus timetable domain; DSTC2 and DSTC3 used human-computer dialogues in the restaurant information domain; DSTC4 dialogues were human-human and in the tourist information domain; DSTC5 also is from the tourist information domain, but training dialogues are provided in one language and test dialogues are in a different language. Finally, as the DSTC6 edition consisted of 3 parallel tracks, different datasets were released for each track, such as, a transaction dialogue dataset for the restaurant domain, two datasets that are part of OpenSubtitles and Twitter datasets, and different chat-oriented dialogue datasets with dialogue breakdown annotations in Japanese and English.

A more recent challenge that started in 2017 and continued into 2018, with its second edition being the Conversational Intelligence Challenge  (ConvAI)\footnote{\url{http://convai.io/}}. This challenge, conducted under the scope of NIPS, has the aim to unify the community around the task of building systems capable of intelligent conversations. In its first edition teams were expected to submit dialogue systems able to carry out intelligent and natural conversations about specific news articles with humans. The aim of the task of the second edition has been to model normal conversation when two interlocutors  meet for the first time, and get to know each other. The dataset of this task consists of 10,981 dialogues with 164,356 utterances, and it is available in the ParlAI framework mentioned above.

Finally, the Alexa Prize\footnote{\url{https://developer.amazon.com/alexaprize}} has attracted mass media and research attention alike. This annual competition for university teams is dedicated at accelerating the field of conversational AI in the framework of the Alexa technology. The participants have to create socialbots that can converse coherently and engagingly with humans on news events and popular topics such as entertainment, sports, politics, technology and fashion. Unfortunately, no datasets have been released.

% !TEX root = survey_dialogue_evaluation.tex
\section{Challenges and Future Trends}
\label{sec:discussion}
In the introduction, we stated  that the goal of the dialogue evaluation is to find methods that are automated, repeatable, are correlated to human judgements, capable of differentiating among various dialogue strategies and explain which features of the dialogue system contribute to its quality. The main motivation behind this is the need to reduce the human evaluation effort as much as possible, since human involvement creates high costs and is highly time-consuming. In this survey, we presented the main concepts regarding evaluation of dialogue systems and showcased the most important methods. However, evaluation of dialogue systems is still an area of open  research. In this section, we summarize the current challenges and future trends that we deem most important.

\paragraph{Automation.} The evaluation methods covered in this survey all achieve a certain degree of automation. However, the automation is achieved with significant engineering effort, or by loss of correlation to human judgements.
Word-overlap metrics (see Section \ref{sec:general_metrics}), which are borrowed from the machine translation and summarization community, are fully automated. However, they do not correlate with human judgements on the turn level. On the other hand, BLEU becomes more competitive when applied on the corpus-level or system-level \citep{galley2015dbleu,Lowe2017AutoTuring}. More recent metrics such as $\Delta$BLEU and ADEM (see Section \ref{sec:general_metrics}) have significantly higher correlations to human judgements while requiring a significant amount of human annotated data as well as thorough engineering.  

Task-oriented dialogue systems can be evaluated semi-automatically or even fully automatically. These systems benefit from having a well-defined task, where success can be measured. Thus, user satisfaction modelling (see Section \ref{sce:user_satisfaction_modelling}) as well as user simulations (see Section \ref{subsec:simulation}) exploit this to automate their evaluation. However, both approaches need a significant amount of engineering and human annotation: user satisfaction modelling usually requires prior annotation effort, which is followed by fitting a model that predicts the judgements. In addition to this effort, the process has to be potentially repeated for each new domain or new functionality that the dialogue system incorporates. Although in some cases the model fitted on the data for one dialogue system can be reused to predict another dialogue system, this is not always possible. 

On the other hand, user simulations require two steps: gathering data to develop a first version of the simulation, and then building the actual user simulation. The first step is only required for user simulations that are based on training corpora (e.g. the neural user simulation). A significant drawback is that the user simulation is only capable of simulating the behaviour which is represented in the corpus or the rules. This means that it cannot cover unseen behaviour well. Furthermore, the user simulation can hardly be used to train or evaluate dialogue systems for other tasks or domains. 

Automation is thus achieved to a certain degree, but with significant drawbacks. Hence, finding ways to facilitate the automation of evaluation methods is clearly an open challenge.

\paragraph{High Quality Dialogues.} One major objective for a dialogue system is to deliver high quality interactions with its users. However, it is often not clear how ``high quality" is defined in this context or how to measure it. For task oriented dialogue systems, the mostly used definition of quality is often measured by means of task success and number of dialogue turns (e.g. a reward of 20 for task-success minus the number of turns needed to achieve the goal). However, this definition is not applicable to conversational dialogue systems and it might ignore other aspects of the interaction (e.g. frustration of the user). Thus, the current trend is to let humans judge the \emph{appropriateness} of the system utterances. However, the notion of appropriateness is highly subjective and entails several finer-grained concepts (e.g. ability to maintain the topic, the coherence of the utterance, the grammatical correctness of the utterance itself, etc.). Currently, appropriateness is modelled by means of latent representations (e.g. ADEM), which are derived again from annotated data. 

Other aspects of quality concern the purpose of the dialogue system in conjunction with the functionality of the system. For instance, \cite{zhou2018design} define the purpose of their conversational dialogue system to build an emotional bond between the dialogue system and the user. This goal differs significantly from the task of training a medical student in the interaction with patients. Both systems need to be evaluated with respect to their particular goal. The ability to build an emotional bond can be evaluated by means of the interaction length (longer interactions are an indicator of a higher user engagement), whereas training (or e-learning) systems are usually evaluated regarding their ability of selecting an appropriate utterance for the given context.

The target audience plays an important role as well. Since quality is mainly a subjective measure, different user groups prefer different types of interactions. For instance, depending on the level of domain knowledge, novice users prefer instructions that use less specialized wording, whereas domain experts might prefer a more specialized vocabulary. 

The notion of quality is thus dependent on a large amount of factors. The evaluation needs to be adapted to take aspects such as the dialogue system's purpose, the target audience, and the dialogue system implementation itself into account. 

\paragraph{Lifelong Learning.} The notion of lifelong learning for machine learning systems has gained traction recently. The main concept of lifelong learning is that a deployed machine learning system continues to improve by interaction with its environment \citep{chen2016lifelong}. Lifelong learning for dialogue systems is motivated by the fact that it is not possible to encounter all possible situations during training, thus, a component that allows the dialogue system to retrain itself and adapt its strategy during deployment seems the most logical solution. 

The evaluation step is critical in order to achieve lifelong learning. Since the dialogue system relies on the ability to automatically find critical dialogue states where it needs assistance, a module is needed which is able to evaluate the ongoing dialogue. One step in this direction is done by \cite{hancock2019learning}, who present a solution that relies on a satisfaction module that is able of to classify the current dialogue state as either satisfactory or not. If this module finds an unsatisfactory dialogue state, a feedback module asks the user for feedback. The feedback data is then used to improve the dialogue system. 

The aspect of lifelong learning brings a large variety of novel challenges. Firstly, the lifelong learning system requires a module that self-monitors its behaviour and notices when a dialogue is going wrong. For this, the module needs to rely on evaluation methods that work automatically, or at least semi-automatically. The second challenge lies in the evaluation of the lifelong learning system itself. The self-monitoring module as well as the adaptive behaviour need to be evaluated. This brings a new dimension of complexity into the evaluation procedure. 

\subsection{Conclusion}
Evaluation is a critical task when developing and researching dialogue systems. Over the past decades, many methods and concepts have been proposed. These methods and concepts are related to the different requirements and functionalities of the dialogue systems. These are subsequently dependent on the current development stage of the dialogue system technology. Currently, the trend is moving towards building end-to-end trainable dialogue systems based on large amounts of data. These systems have different requirements for evaluation than a finite state, machine-based system. Thus, the problem of evaluation is evolving in tandem to the progress of the dialogue system technology itself. This survey presents the current state-of-the-art research in evaluation.

\section*{Acknowledgements}
We would like to thank Lina Scarborough for proofreading the manuscript. We would also like to thank our anonymous reviewers for their valuable reviews that helped improve the quality of this survey. 

\section*{Funding}
This work was supported by the LIHLITH project in the framework of EU ERA-Net CHIST-ERA; the Swiss National Science Foundation [20CH21\_174237]; the Spanish Research Agency [PCIN-2017-11, PCIN-2017-085/AEI]; Eneko Agirre and Arantxa Otegi received the support of the UPV/EHU [grant GIU16/16]; Agence Nationale pour la Recherche [ANR-17-CHR2-0001-03]
\section*{Conflicts of Interest}
There are no conflicts of interest to disclose.

\bibliographystyle{spbasic}  
\bibliography{survey_dialogue_evaluation} % Put here the name of BibTeX file

\end{document}